\documentclass[preprint,12pt]{elsarticle}
\usepackage{amssymb}
\usepackage{graphicx}
\usepackage{amsmath}
\usepackage{longtable}
\usepackage{algorithm} 
\usepackage{algpseudocode} 
\usepackage{mathrsfs}
\usepackage{subcaption}

\usepackage{mathtools}
\usepackage{pifont}
\usepackage{color}
\usepackage{lineno}
\usepackage{makecell} 
\usepackage{pdflscape}
\usepackage{adjustbox}
\usepackage[utf8]{inputenc}
\usepackage{tabularx}
\usepackage{blindtext}
\usepackage{setspace}
\usepackage{notoccite} 
\usepackage{lscape} 
\usepackage{caption} 
\usepackage{mwe}
\usepackage{booktabs}
\usepackage{hyperref}
\usepackage{amsthm}

\newcommand{\RNum}[1]{\lowercase\expandafter{\romannumeral #1\relax}}
\newcommand{\RNumU}[1]{\uppercase\expandafter{\romannumeral #1\relax}}
\usepackage{natbib}
\journal{Elsevier}

\begin{document}
\begin{frontmatter}
\title{HawkEye: Advancing Robust Regression with Bounded, Smooth, and Insensitive Loss Function}
\author[inst1]{Mushir Akhtar}
\author[inst1]{M. Tanveer\corref{Correspondingauthor}}
\author[inst1]{Mohd. Arshad}
\affiliation[inst1]{organization={Department of Mathematics, Indian Institute of Technology Indore},
            addressline={Simrol}, 
            city={Indore},
            postcode={453552}, 
            state={Madhya Pradesh},
            country={India}}
            \cortext[Correspondingauthor]{Corresponding author}
\begin{abstract}
Support vector regression (SVR) has garnered significant popularity over the past two decades owing to its wide range of applications across various fields. Despite its versatility, SVR encounters challenges when confronted with outliers and noise, primarily due to the use of the $\varepsilon$-insensitive loss function. To address this limitation, SVR with bounded loss functions has emerged as an appealing alternative, offering enhanced generalization performance and robustness. Notably, recent developments focus on designing bounded loss functions with smooth characteristics, facilitating the adoption of gradient-based optimization algorithms. However, it's crucial to highlight that these bounded and smooth loss functions do not possess an insensitive zone. In this paper, we address the aforementioned constraints by introducing a novel symmetric loss function named the HawkEye loss function. It is worth noting that the HawkEye loss function stands out as the first loss function in SVR literature to be bounded, smooth, and simultaneously possess an insensitive zone. Leveraging this breakthrough, we integrate the HawkEye loss function into the least squares framework of SVR and yield a new fast and robust model termed HE-LSSVR. The optimization problem inherent to HE-LSSVR is addressed by harnessing the adaptive moment estimation (Adam) algorithm, known for its adaptive learning rate and efficacy in handling large-scale problems. To our knowledge, this is the first time Adam has been employed to solve an SVR problem. To empirically validate the proposed HE-LSSVR model, we evaluate it on UCI, synthetic, and time series datasets. The experimental outcomes unequivocally reveal the superiority of the HE-LSSVR model both in terms of its remarkable generalization performance and its efficiency in training time.


\end{abstract}

\begin{keyword}
Supervised learning \sep Support vector regression \sep Loss function  \sep HawkEye loss function \sep Adam algorithm
\end{keyword}
\end{frontmatter}

\section{Introduction}
Machine learning, a subset of artificial intelligence, has witnessed remarkable advancements in recent years, revolutionizing various fields through its ability to extract meaningful patterns from complex data \cite{goldenberg2019new}. Within the realm of machine learning, supervised learning stands as a cornerstone, where algorithms learn from labeled datasets to make predictions or infer relationships between input and output variables \cite{hastie2009overview}.
Among the diverse applications of supervised learning, regression emerges as a fundamental subfield dedicated to modeling the intricate connections between input variables and continuous output values \cite{angarita2019taxonomy}. \\
Support vector regression (SVR) \cite{drucker1996support} is a powerful and versatile machine learning technique designed for solving regression problems. It belongs to the family of support vector machines (SVMs) \cite{cortes1995support}, which are renowned for their effectiveness in classification tasks. SVMs have applications in several areas, such as pattern classification \cite{dong2022support}, concept drift \cite{galmeanu2022weighted}, credit risk evaluation \cite{TANG2021327}, and many more. In a nutshell, SVMs are rooted in the framework of statistical learning theory, which equips them with the ability to generalize effectively to new and unseen data. However, SVR extends the capabilities of SVM to handle continuous output variables, making it a valuable tool for predictive modeling and function approximation. It has extensive applications across various domains, such as
load forecasting \cite{fan2021forecasting}, water temperature prediction \cite{quan2022research}, stock prediction \cite{dash2023fine}, landslide prediction \cite{ma2022metaheuristic}, petroleum resource prediction \cite{wang2023novel}, health estimation of battery \cite{li2022state}, estimation of bank profitability \cite{zahariev2022estimation}, flight control \cite{shin2011adaptive} and so forth. 
\par
Consider a regression training dataset $\left\{x_i,y_i\right\}_{i=1}^N$, where $x_i \in \mathbb{R}^m$ represents the input data point and $y_i \in\mathbb{R}$ corresponds to the target value associated with that input.
The core idea of SVR is to estimate a linear function $f(x)= w^\intercal x+b$, where $b \in \mathbb{R}$ represents the bias and $w \in \mathbb{R}^m$ signifies the weight vector. These parameters are estimated through the process of training using available data. For a test data point $\widetilde{x}$, the associated target value $\widetilde{y}$ is estimated as $w^\intercal \widetilde{x}+b$. Analogously to SVM, we discuss two scenarios of SVR namely, hard SVR and soft SVR. For pedagogical reasons, we begin by describing the case of hard SVR. In hard SVR, the goal is to find a linear regression function $f$ that fits the training data while strictly enforcing that all data points are within a specified margin of error $\varepsilon$. Mathematically, the objective function for hard SVR is as follows:
\begin{align} \label{hardmarginSVR}
\underset{w, b}{min} \hspace{0.2cm} &\frac{1}{2}\|w\|^2 \nonumber \\
 \text { subject to }\hspace{0.2cm}  & y_i-w^\intercal x_i-b \leq \varepsilon \\
 &w^\intercal x_i+b-y_i \leq \varepsilon, ~\forall~ i=1,2, \ldots,N. \nonumber
\end{align}
The two constraints ensure that the predicted values for all data points must lie within a symmetric margin of tolerance $\varepsilon$ around the true target values $y_i$. In other words, the error between the predicted values and the true target values is required to be no more than $\varepsilon$. The tacit assumption in (\ref{hardmarginSVR}) is that such a function $f$ actually
exists that approximates all pairs $\left(x_i, y_i\right)$ with $\varepsilon$ precision. However, this may not be the case in real-world applications. To address this challenge, a widely adopted strategy is to permit a certain degree of error in the model while simultaneously imposing penalties for these errors. This is accomplished by introducing a loss function into the objective function and called as soft SVR. The unconstrained optimization problem of soft SVR is as follows:
\begin{align}
\underset{w, b}{min} \hspace{0.2cm}& \frac{1}{2}\|w\|^2+ C \sum_{i=1}^N \mathcal{L}\left(r_i\right),
\end{align}
where $r_i= y_i - f(x_i)$ represents the training error, $C>0$ is a regularization parameter which controls the trade-off between model complexity and training errors, $\mathcal{L}(\cdot)$ is a loss function, and $\sum_{i=1}^N \mathcal{L}\left(r_i\right)$ expresses the empirical risk.
In simpler terms, soft SVR prioritizes minimizing errors as long as they remain within the $\varepsilon$ tolerance range, but it strongly penalizes any deviations exceeding this threshold.
\par
In this paper, our primary focus revolves around enhancing the performance of SVR. In the realm of developing novel regression models, one critical aspect that garners significant attention is the design of appropriate loss functions.
The traditional SVR models commonly employ convex and unbounded loss functions, such as the $\varepsilon$-insensitive loss function \cite{drucker1996support}, the least squares loss function \cite{suykens1999least}, and the Huber loss function \cite{muller1997predicting}, to penalize regression errors. These loss functions entail that as the error between predicted and true values increases, the corresponding loss also increases without bounds. This characteristic can pose several challenges, particularly when dealing with noisy data, and may result in suboptimal model performance.
\par
In real-world applications, it may possible that data contains outliers. The term ``outliers" appertains to instances that deviates significantly from the pattern established by the vast majority of the data \cite{hawkins1980identification}. Outliers are oftenly difficult to recognize in high dimensional datasets by reason of the curse of dimensionality \cite{steinwart2008support}. As a result, the learned model is unreliable. To mitigate the effect of outliers, many robust loss functions have been proposed. Over the last few decades, non-convex and bounded loss functions have gained prominence, primarily to enhance the robustness of models in the presence of noisy data. One well-known representative of such functions is the ramp loss function \cite{collobert2006trading}. Its primary purpose is to restrict the maximum loss to a fixed value for data samples that exhibit significant deviations, which diminishes the effect of noise and outliers. Following the introduction of the ramp loss function, several variants have been developed, further extending the versatility of this approach. These variants include the non-convex least square loss function \cite{wang2014robust}, ramp $\varepsilon$-insensitive loss function \cite{liu2015ramp}, generalized quantile loss function \cite{yang2019robust}, quadratic non-convex $\varepsilon$-insensitive loss function \cite{ye2020robust} and canal loss function \cite{liang2022kernel}. The key advantage of these loss functions lies in their bounded nature. They effectively constrain the loss to a predefined threshold, ensuring that the model remains robust even in the presence of noisy data. However, it is essential to recognize that the methodology of hard truncation employed in these loss functions, especially when addressing data points with substantial regression errors, can introduce non-smoothness into the loss function. Hence, the optimization problems associated with SVR models based on non-convex and non-smooth loss functions present algorithmic intricacies and demand considerable computational resources.
\par
Based on insights gleaned from previous research efforts, this paper introduces the HawkEye loss, an innovative loss function meticulously designed to exhibit boundedness, smoothness, and the simultaneous presence of an insensitive zone. The key advantage of the smoothness property is that it allows the utilization of gradient-based optimization algorithms, ensuring well-defined gradients and enabling the application of fast and reliable optimization techniques. Subsequently, by 
incorporating the newly proposed HawkEye loss into the least squares version of SVR, we propose a new fast and robust model for handling large-scale problems called HE-LSSVR. The optimization problem associated with the HE-LSSVR is effectively addressed through the utilization of the adaptive moment estimation (Adam) algorithm. The key contributions of this study can be succinctly summarized as follows:
\begin{itemize}
    \item We introduce a novel development in the realm of supervised learning, the HawkEye loss function, meticulously crafted to exhibit boundedness, smoothness, and the simultaneous inclusion of an insensitive zone.
    \item We present a comparative analysis of the existing loss functions with the HawkEye loss function and showcase that it is the first loss function in SVR literature that is bounded, smooth, and simultaneously possesses an insensitive zone.
    \item We amalgamate the HawkEye loss function into the least squares framework of SVR and formulate a new fast and robust model coined as HE-LSSVR.
    \item We address the optimization problem of HE-LSSVR by harnessing the Adam algorithm, known for its adaptive learning rate and efficacy in handling large-scale problems. To our knowledge, this is the first time Adam has been employed to solve an SVR problem.
    \item We conduct numerical experiments on UCI, synthetic, and time series datasets and validate that the proposed HE-LSSVR model is superior compared to the baseline models both in terms of generalization performance and training time.
\end{itemize}

\section{Related work}
\begin{figure*}
\centering
    \subcaptionbox{     \label{fig:least square loss}} { %
      \includegraphics[width=0.48\textwidth,keepaspectratio]{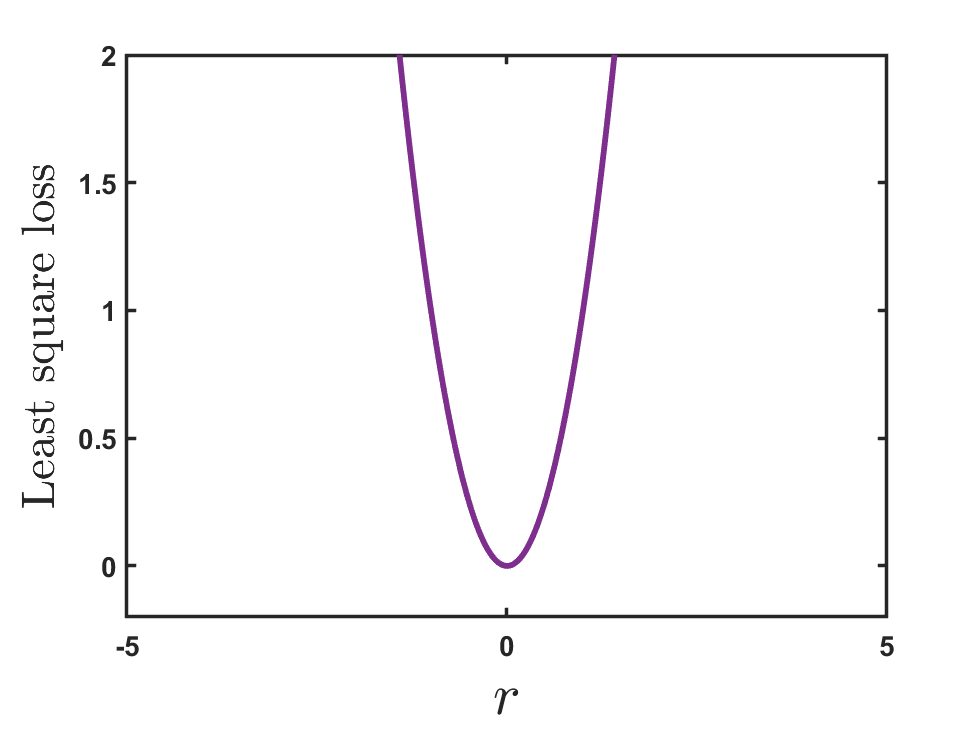}}
      \hfill
      \subcaptionbox{   \label{fig:epsilon insensitive}} { %
      \includegraphics[width=0.48\textwidth,keepaspectratio]{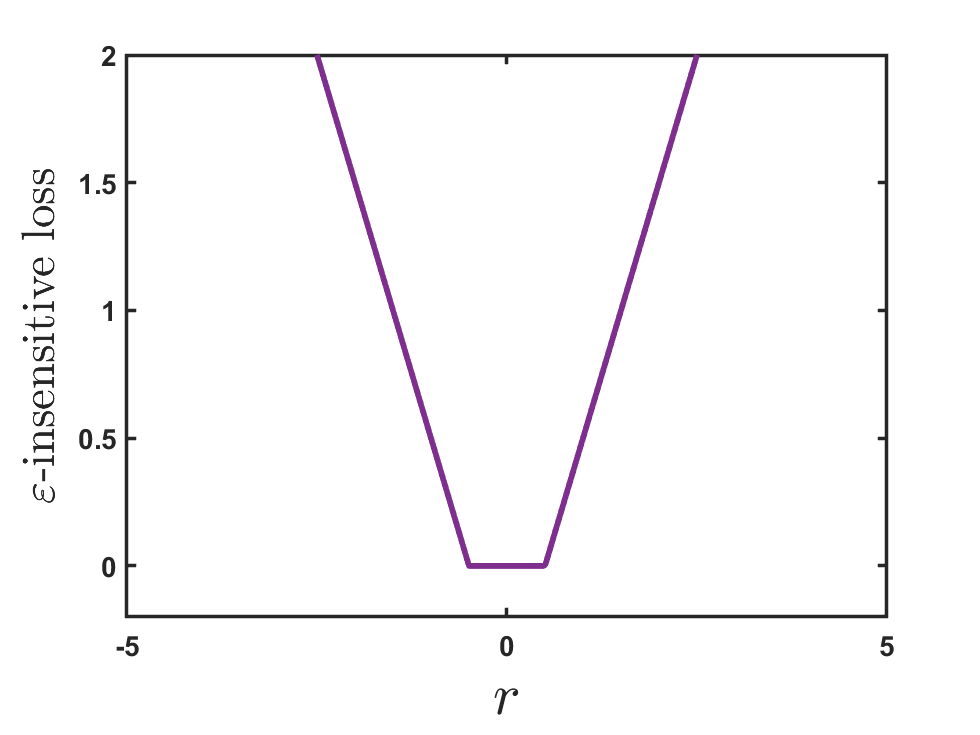}}
\\
      \subcaptionbox{  \label{fig:huber loss}} { %
      \includegraphics[width=0.48\textwidth,keepaspectratio]{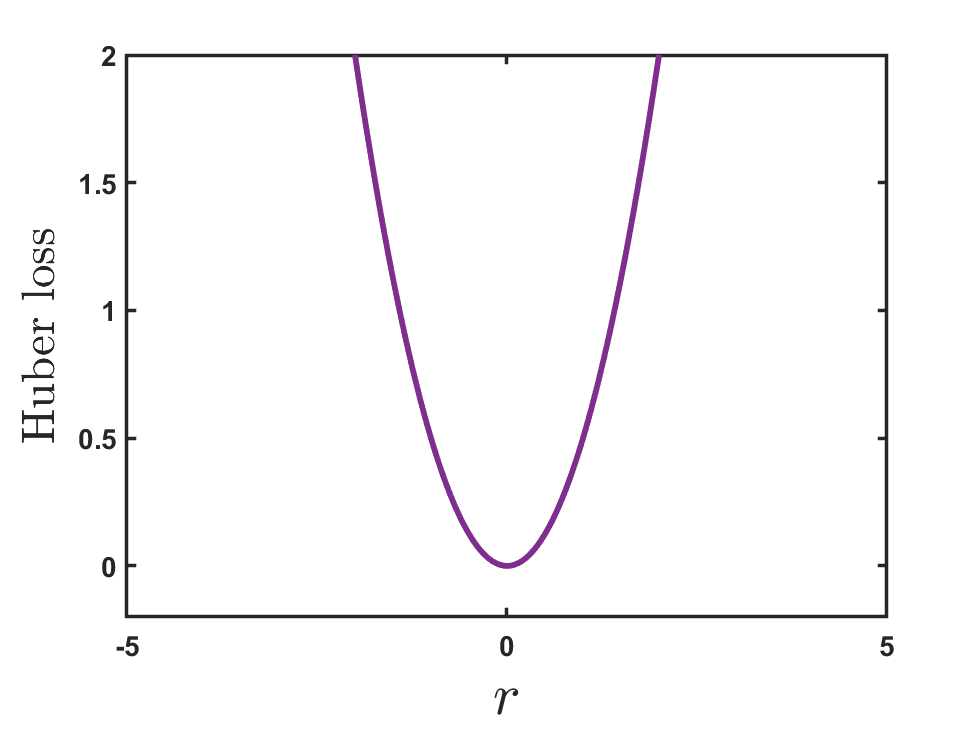}}
      \hfill
      \subcaptionbox{  \label{fig:canal loss}} { %
      \includegraphics[width=0.48\textwidth,keepaspectratio]{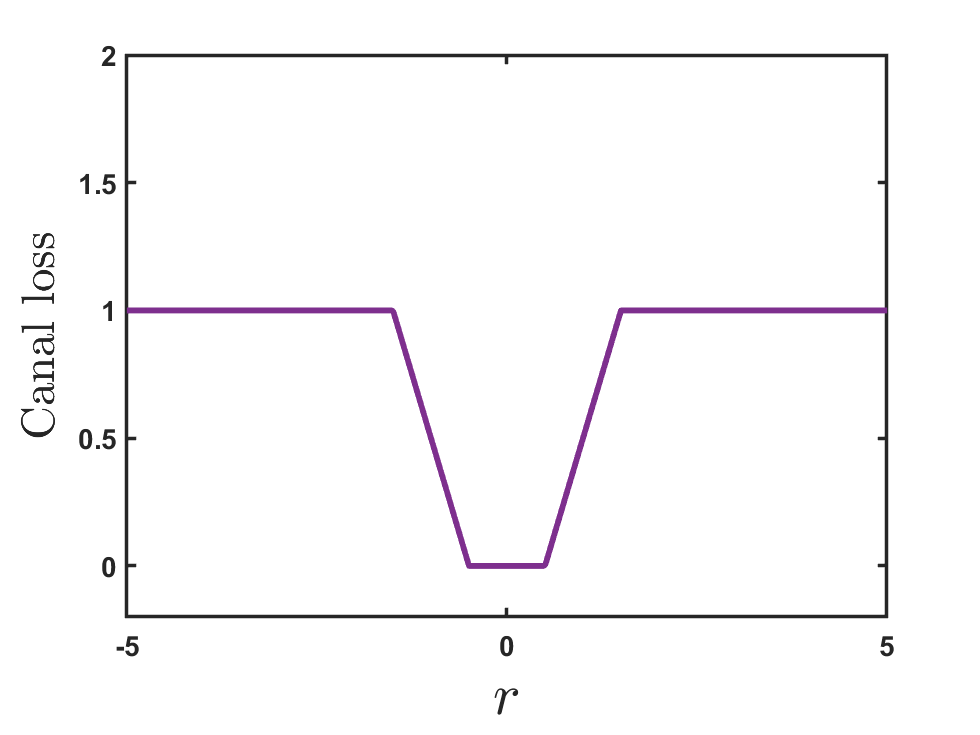}}
\\
    \subcaptionbox{     \label{fig:ramp insensitive }} { %
      \includegraphics[width=0.48\textwidth,keepaspectratio]{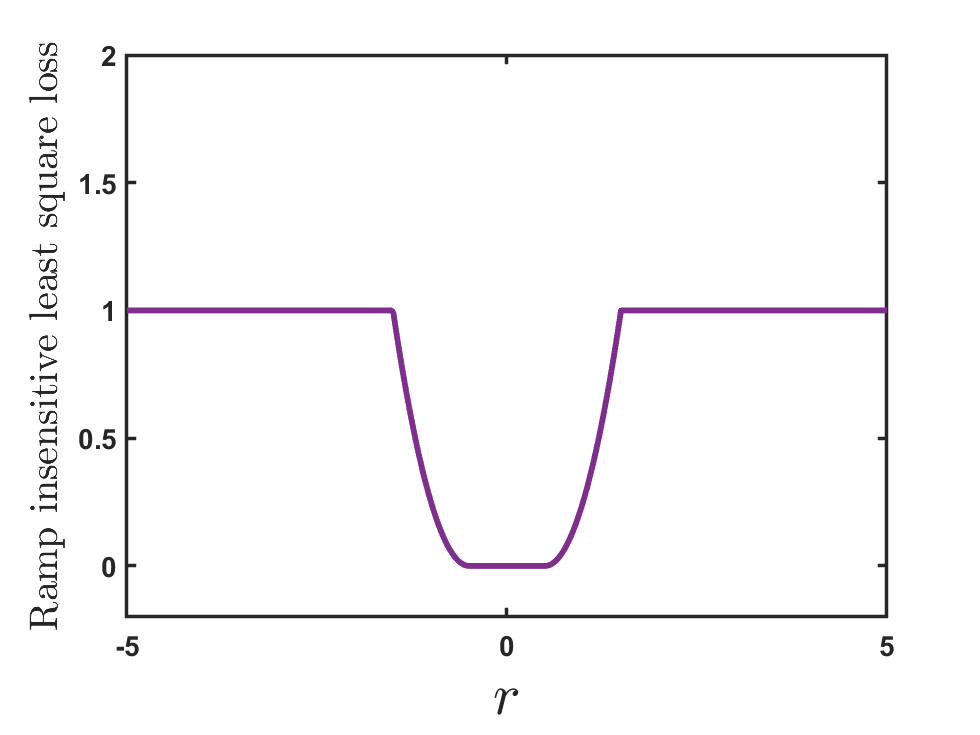}}
      \hfill
      \subcaptionbox{   \label{fig:bounded square }} { %
      \includegraphics[width=0.48\textwidth,keepaspectratio]{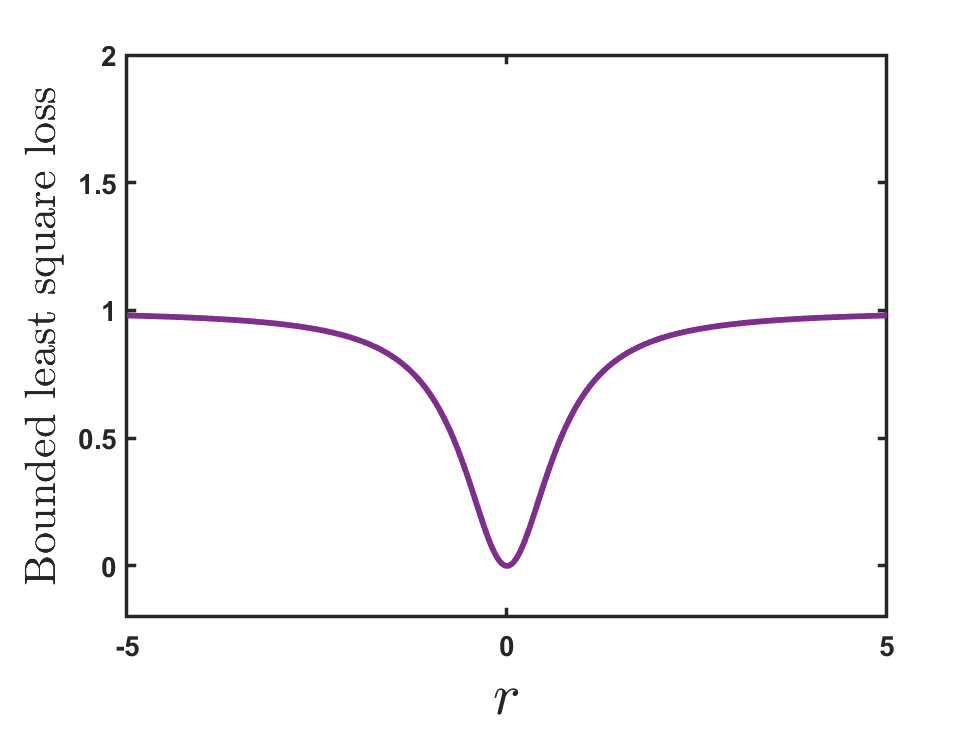}}
      \caption{Visual illustration of some baseline loss functions. (a) Least square loss function. (b) $\varepsilon$-insensitive loss function with $\varepsilon=0.5$. (c) Huber loss function with $\theta=5$. (d) Canal loss function with $\varepsilon=0.5$ and $\theta=1.5$. (e) Ramp insensitive least square loss function with $\varepsilon=0.5$ and $\theta=1.5$. (f) Bounded least square loss function with $t=1$ and $\theta=2$.}
    \label{fig:Baseline Loss functions}
 \end{figure*}

The design of loss functions is a fundamental component in the development of machine learning models \cite{barron2019general}. It plays a pivotal role in assessing the dissimilarity between predicted values and their corresponding target values \cite{frogner2015learning}. In this section, we provide a brief overview of commonly employed loss functions in regression.
To facilitate comprehension, we present the mathematical formulation of these loss functions in Table \ref{tab:baseline_loss-table}, and offer graphical representations of some of them in Figure \ref{fig:Baseline Loss functions}.

\begin{table}[]
\caption{Mathematical formulation of commonly used loss function in regression. Unless otherwise specified $\varepsilon$, $\theta$, and $t$ are non negative hyperparameters.}
\label{tab:baseline_loss-table}
\resizebox{\textwidth}{!}{%
\begin{tabular}{lll}
\hline
Category & Loss function & Formulation \\ \hline
 &  &  \\
 & Least square loss function & $\mathcal{L}\left(r\right) = r^2$ \\
 &  &  \\
 & Absolute loss function & $\mathcal{L}\left(r\right) = |r|$  \\
Unbounded &  &  \\
 & Huber loss function & 
$\mathcal{L}\left(r\right) =
\begin{cases}
\theta |r|- \frac{1}{2}\theta^2, & |r| \geq \theta \\
\frac{1}{2}r^2, & |r| < \theta 
\end{cases}
$
 \\
 &  &  \\
 & Insensitive loss function & $\mathcal{L}\left(r\right) = \underset{}{max} \hspace{0.2cm} \{ 0, |r| - \varepsilon \}$ \\
 &  &  \\ \hline
 & Ramp insensitive loss function & $\mathcal{L}\left(r\right) = \begin{cases}
0, & |r| < \varepsilon \\
|r|-\varepsilon, & \varepsilon \leq |r| \leq \theta\\
\theta - \varepsilon, & |r| > \theta
\end{cases}
$ \\
 &  &  \\
 & Non-convex least square loss function & $\mathcal{L}\left(r\right) =
\begin{cases}
 r^2, & |r| \leq \theta \\
\theta^2, & |r| > \theta 
\end{cases}
$  \\
 &  &  \\
 & Ramp insensitive least square loss function &  $\mathcal{L}\left(r\right) = \begin{cases}
0, & |r| < \varepsilon \\
(|r|-\varepsilon)^2, & \varepsilon \leq |r| \leq \theta\\
(\theta - \varepsilon)^2, & |r| > \theta
\end{cases}
$  \\ Bounded &  &  \\
 & Quadratic non-convex insensitive loss function & $\mathcal{L}\left(r\right) = \begin{cases}
0, & |r| < \varepsilon \\
(|r|-\varepsilon)^2, & \varepsilon \leq |r| \leq t\\
(t - \varepsilon)^2 + \theta |r| - \theta t, & |r| > t
\end{cases}
$  \\
 &  &  \\
 & Canal loss &  $\mathcal{L}\left(r\right) = \underset{}{min} \hspace{0.2cm} \{ \theta - \varepsilon, \underset{}{max} \hspace{0.1cm} \{ 0, |r| - \varepsilon \}    \}$ \\
 &  &  \\
 & Bounded  least square loss function & $\mathcal{L}\left(r\right) =  \frac{1}{t} \left( 1- \frac{1}{1+ \theta r^2}  \right)$ \\
 &  &  \\ \hline
\end{tabular}%
}
\end{table}

\begin{subsection}{Unbounded loss functions}
The least squares loss function stands as the dominant choice in the realm of regression, characterized by its smoothness and quadratic growth as the regression error escalates. Its widespread application spans across diverse domains \cite{li2019multi,xie2020general}, with least squares support vector regression (LS-SVR) \cite{suykens1999least} representing a notable and successful adaptation. However, its rapid growth rate renders it sensitive to the presence of noise in the data \cite{zhong2012training}. In contrast to least squares loss, the absolute loss function \cite{narula1982minimum} and pinball loss function \cite{anand2020new} exhibit linear growth as errors increase and are robust against noise. Nevertheless, these loss functions lack the desirable smoothness property, which can pose challenges for optimization and may lead to slower convergence during model training. Encouragingly, the Huber loss function \cite{muller1997predicting} offers a promising middle ground. It amalgamates the attributes of least squares and absolute loss functions, making it smooth and featuring a relatively gentle growth rate. This characteristic allows it to impose linear penalties on large errors and quadratic penalties on smaller errors. The Huber loss function is widely used due to its balance between robustness and smoothness. Furthermore, numerous extensions of the Huber loss function \cite{balasundaram2019robust,gupta2021regularization} have been developed, extending its versatility in various applications. In the aforementioned loss functions, all training data points contribute to the regression, often leading to models with reduced sparsity. In contrast, the $\varepsilon$-insensitive loss function takes a different approach by not penalizing data points that fall within an $\varepsilon$-tube around the target values. This offers an element of flexibility and sparsity in the model. While these diverse loss functions have demonstrated significant success in regression, it's imperative to acknowledge their inherent unbounded nature, which renders the corresponding regressors susceptible to the influence of noise and outliers with substantial deviations.

\begin{subsection}{Bounded loss functions}
In the pursuit of creating more robust models capable of withstanding the impact of noise, several bounded loss functions have been proposed in the literature. One prominent example is the ramp loss function, originally developed for classification and later adapted for regression \cite{collobert2006trading}. It imposes a strict limit on the maximum allowable loss for data points with substantial deviations. This approach effectively prevents noise and outliers from exerting an excessive influence, contributing to the model's robustness. Various variations of the ramp loss function, tailored for regression, have been proposed in the literature. The ramp $\varepsilon$-insensitive loss function \cite{tang2018ramp} is a notable example in this category. It assigns zero penalties to data points with minor errors, while significant errors are met with finite penalties. This well-calibrated design achieves a delicate equilibrium between enhancing robustness and preserving model sparsity. To further emphasize the penalty applied to non-outlier observation errors, the non-convex least squares loss function \cite{wang2014robust} and the ramp $\varepsilon$-insensitive least squares loss function \cite{liu2016ramp} are introduced in the literature. These incarnations replace the linearly increasing segment of the ramp loss function with a squared form. In recent developements, \citet{ye2020robust} and \citet{liang2022kernel} respectively proposed the quadratic non-convex insensitive loss function and the canal loss function which are also tailored to address the challenges posed by noisy labels by enforcing a fixed loss on data points with notable deviations. However, it's crucial to acknowledge that most of the aforementioned discussed loss functions adopt the hard truncation strategy which imparts a non-smooth character to these loss functions.
\end{subsection}

\end{subsection}
\section{Proposed work}
In this section, we propose a novel loss function named HawkEye loss, and leveraging this breakthrough, we incorporate the proposed HawkEye loss function into the least squares framework of SVR and yield a new fast and robust support vector regression model coined as HE-LSSVR.

\subsection{HawkEye loss function}
\begin{figure*}
\centering
    \subcaptionbox{     \label{fig:first}} { %
      \includegraphics[width=0.48\textwidth,keepaspectratio]{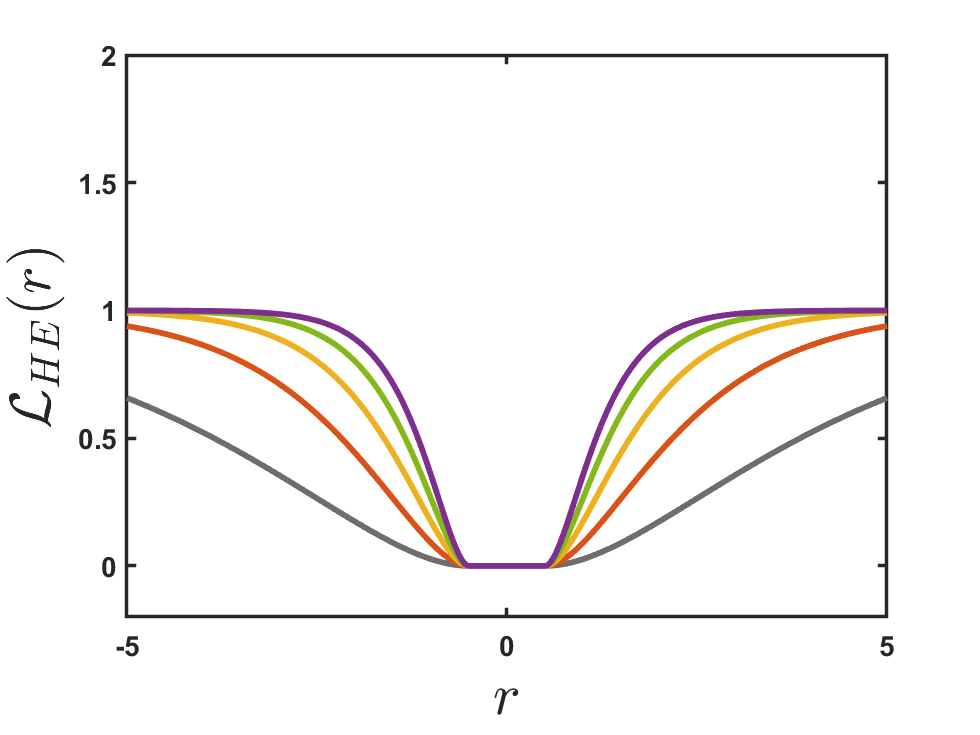}}
      \hfill
      \subcaptionbox{   \label{fig:second}} { %
      \includegraphics[width=0.48\textwidth,keepaspectratio]{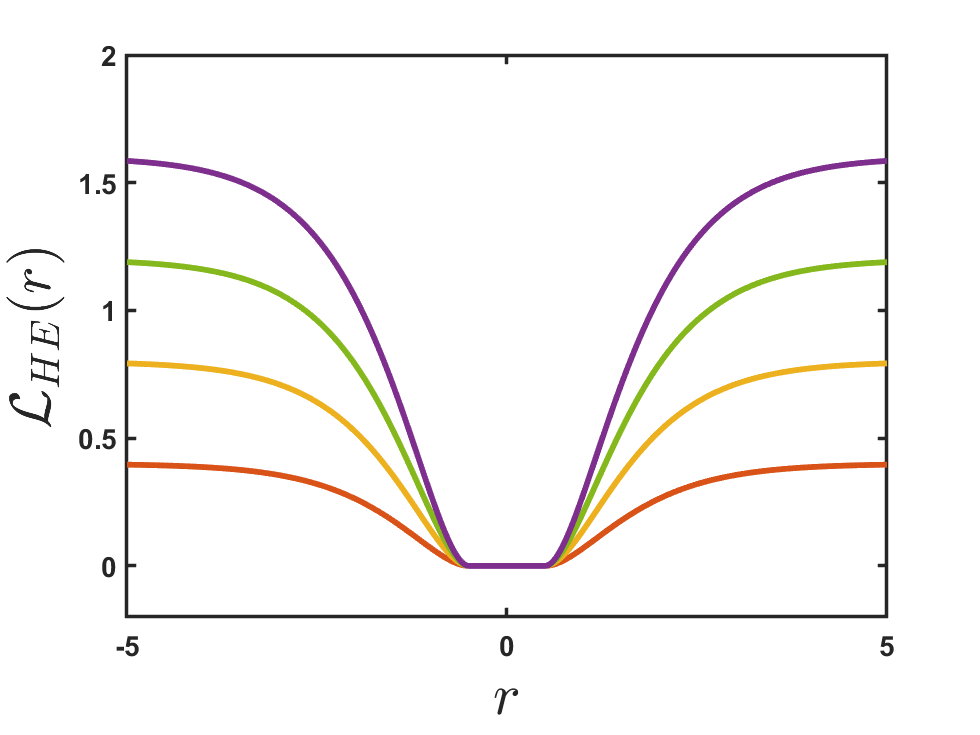}}
\\
      \subcaptionbox{  \label{fig:third}} { %
      \includegraphics[width=0.48\textwidth,keepaspectratio]{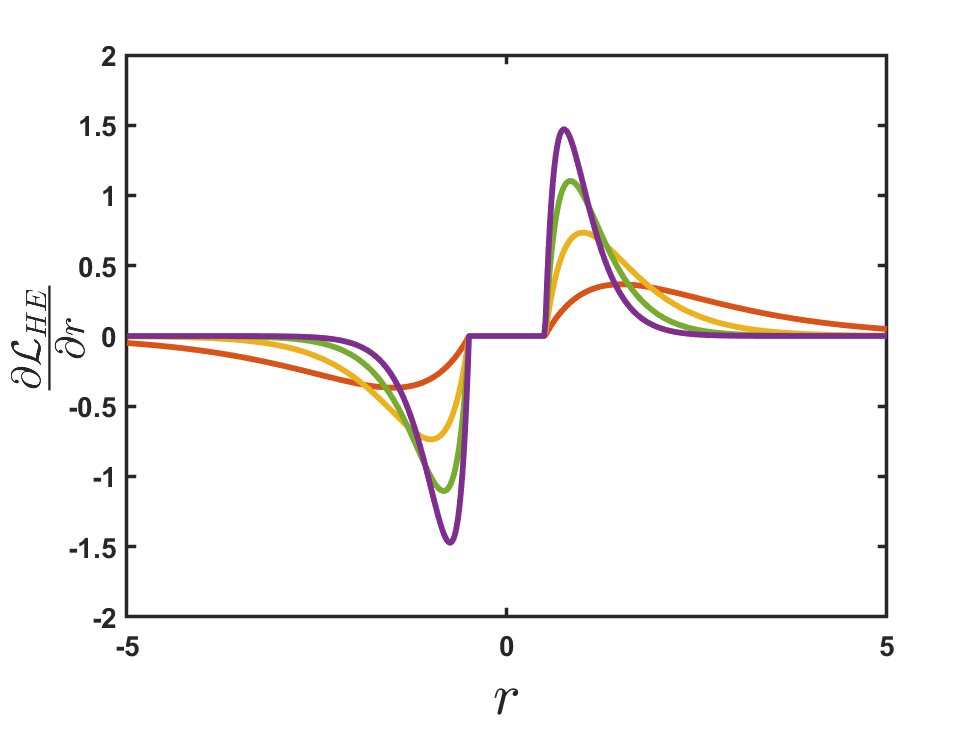}}
      \hfill
      \subcaptionbox{  \label{fig:four}} { %
      \includegraphics[width=0.48\textwidth,keepaspectratio]{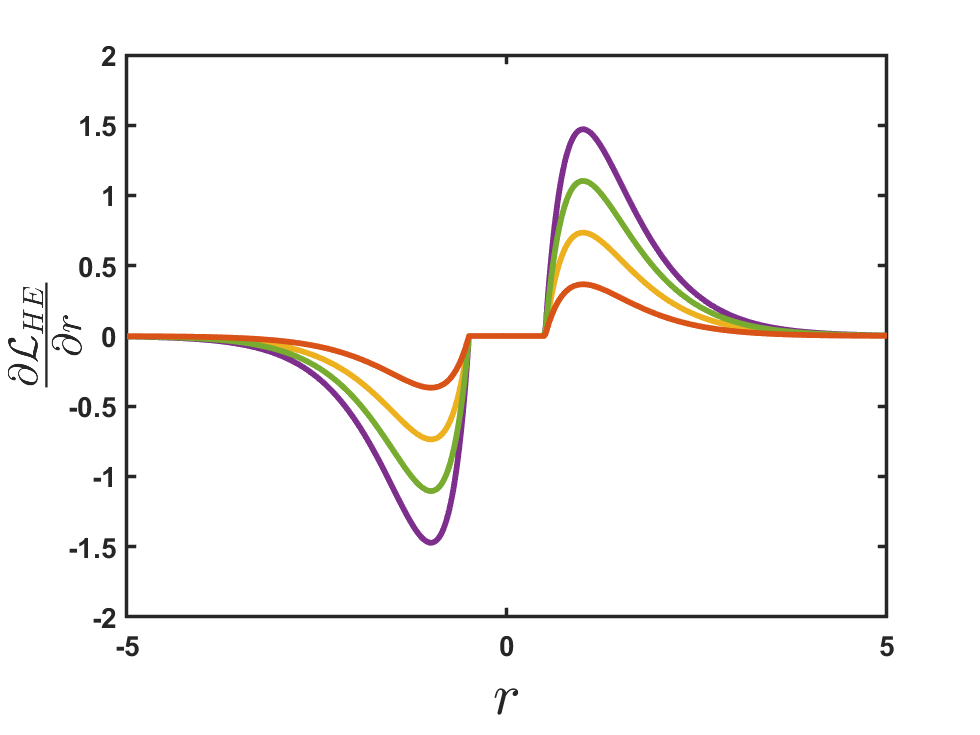}}
      \caption{The proposed HawkEye loss function and its gradient for different values of $a$ and $\lambda$. Subfigures (a), (c) are plotted for fixed $\lambda$ and different values of $a$ while (b), (d) are plotted for fixed $a$ and different values of $\lambda$.
      }
    \label{fig:Proposed Loss and its Gradient}
 \end{figure*}

We present a noteworthy breakthrough in the field of supervised learning: a novel symmetric loss function, termed the HawkEye loss function. It is designed to manifest desirable attributes of boundedness and smoothness while concurrently incorporating an insensitive zone. This innovative approach marks a substantial advancement in optimizing the training process of machine learning models. The mathematical expression of the proposed HawkEye loss function is articulated as follows:
\begin{align}\label{Proposed Loss}
\mathcal{L}_{HE}(r,a,\lambda)=
\begin{cases}
\lambda \left[  1-\{-a(r+\varepsilon)+1\} e^{a(r+\varepsilon)}   \right], & r \leq -\varepsilon, \\
0, & -\varepsilon < r < \varepsilon,\\
\lambda \left[  1-\{a(r-\varepsilon)+1\} e^{-a(r-\varepsilon)}   \right], & r \geq \varepsilon,
\end{cases}
\end{align}
where $\varepsilon > 0$ is an insensitivity parameter that serves to establish a zone of tolerance within the loss function, $a \in \mathbb{R}^+$ is a shape parameter that determines the curvature and steepness of the loss function, and $\lambda \in \mathbb{R}^+$ is a bounding parameter that acts as a constraint to ensure that the loss does not exceed a specified bound. To visually depict the proposed HawkEye loss function (\ref{Proposed Loss}), Figures \ref{fig:first} and \ref{fig:second} showcase its representation for various combinations of the parameters $a$ and $\lambda$. The Proposed HawkEye loss function (\ref{Proposed Loss}), as delineated in this work, exhibits the following intrinsic properties:\\

\noindent
\textbf{Property 1}:  $\mathcal{L}_{HE}(\cdot)$ exhibits sparsity, meaning it assigns zero loss to data points falling within a specified $\varepsilon$-insensitive zone.
\begin{proof}   
For $-\varepsilon < r < \varepsilon$, the loss function $\mathcal{L}_{HE}(\cdot)$ is defined as
$\mathcal{L}_{HE}(r) = 0$.
This demonstrates that within the $\varepsilon$-insensitive zone, the loss is indeed zero, proving the sparsity property.
\end{proof}
\noindent
\textbf{Property 2}: $\mathcal{L}_{HE}(\cdot)$ is symmetric, signifying its impartial allocation of equal loss to deviations on both sides of the reference point.
\begin{proof}
Consider two values $r$ and $-r$. From (\ref{Proposed Loss})  ,we can easily show that:
$$\mathcal{L}_{HE}(r) = \mathcal{L}_{HE}(-r).$$
This symmetry property is a direct consequence of the construction of the loss function.
\end{proof}
\noindent
\textbf{Property 3}: $\mathcal{L}_{HE}(\cdot)$ is non-negative, ensuring that the loss is always greater than or equal to zero.
\begin{proof}
For all values of $r$, the loss function $\mathcal{L}_{HE}(\cdot)$ is defined as a combination of terms that are non-negative:
\begin{align*}
&\mathcal{L}_{HE}(r) = \lambda \left[  1-\{-a(r+\varepsilon)+1\} e^{a(r+\varepsilon)}   \right], r \leq -\varepsilon,\\
&\mathcal{L}_{HE}(r) = 0, -\varepsilon < r < \varepsilon,\\
&\mathcal{L}_{HE}(r) = \lambda \left[  1-\{a(r-\varepsilon)+1\} e^{-a(r-\varepsilon)}   \right],  r \geq \varepsilon.
\end{align*}
It can be seen easily that all the terms in the definition of $\mathcal{L}_{HE}(\cdot)$ are non-negative, ensuring the non-negativity property. 
\end{proof}
\noindent
\textbf{Property 4}: $\mathcal{L}_{HE}(\cdot)$ is zero at origin and increases monotonically with respect to $|r|$, i.e., $\mathcal{L}_{HE}(0)=0$ and $\frac{\partial\mathcal{L}_{HE}}{\partial |r|} \geq 0.$
\begin{proof}
Since $\varepsilon >0$, the origin will always lie in the region $-\varepsilon< r <\varepsilon$. Then, from (\ref{Proposed Loss}), we can easily see that $\mathcal{L}_{HE}(0)=0$. Now, to prove that the HawkEye loss function increases monotonically with respect to $|r|$, let’s analyze each region separately.\\
For $r \leq -\varepsilon$:\\
We can write
$\mathcal{L}_{HE}(|r|) = \lambda \left[  1-\{-a(-|r|+\varepsilon)+1\} e^{a(-|r|+\varepsilon)}   \right]$.\\
Taking the derivative with respect to $|r|$, we get: \\
$$\frac{\partial\mathcal{L}_{HE}}{\partial |r|}= -\lambda a^2 (-|r|+\varepsilon) e^{a(-|r|+\varepsilon)}.$$
Since $\lambda$, $a >0$, $-|r|+\varepsilon \leq 0$, and $e^{a(-|r|+\varepsilon)}$ is always positive, then we can surely say that the derivative is always non-negative.\\
\noindent
For $-\varepsilon < r < \varepsilon$:\\
In this region, the HawkEye loss function is constant, so $\frac{\partial\mathcal{L}_{HE}}{\partial |r|}=0$.\\
\noindent
For $r \geq \varepsilon$:\\
We can write
$\mathcal{L}_{HE}(|r|) = \lambda \left[  1-\{a(|r|-\varepsilon)+1\} e^{-a(|r|-\varepsilon)}   \right]$.\\
Taking the derivative with respect to $|r|$, we get: \\
$$\frac{\partial\mathcal{L}_{HE}}{\partial |r|}= \lambda a^2 (|r|-\varepsilon) e^{-a(|r|-\varepsilon)}.$$
Since $\lambda$, $a >0$, $|r|-\varepsilon \geq 0$, and $e^{-a(|r|-\varepsilon)}$ is always positive, then the derivative is always non-negative.\\
In conclusion, the HawkEye loss function increases monotonically with respect to $|r|$, i.e., $\frac{\partial\mathcal{L}_{HE}}{\partial |r|} \geq 0.$ 

\end{proof}
\noindent
\textbf{Property 5}: $\mathcal{L}_{HE}(\cdot)$ is bounded by $\lambda$, ensuring that the loss does not grow unbounded as the error increases, thereby enhancing robustness.
\begin{proof}
For this, we analyze its behavior in the three defined regions.\\
For $r \leq -\varepsilon$: \\
In this region, we have:
\begin{align}\label{iii}
\mathcal{L}_{HE}(r) = \lambda \left[  1-\{-a(r+\varepsilon)+1\} e^{a(r+\varepsilon)}   \right].
\end{align}
To prove that $\mathcal{L}_{HE}(r)$ is bounded by $\lambda$, we need to show that the expression in the square bracket of equation (\ref{iii}) is less than or equal to $1$, i.e., $\{-a(r+\varepsilon)+1\} e^{a(r+\varepsilon)} \geq 0$.\\
Let's simplify the expression. \\
$e^{a(r+\varepsilon)}$ is a positive term. \\
$-a(r + \varepsilon) + 1 \geq 0$ (since $a > 0$ and $r \leq -\varepsilon$).\\
Thus, the expression in the square bracket of equation (\ref{iii}) is less than or equal to $1$. Therefore, in this region, $\mathcal{L}_{HE}(r)$ is bounded by $\lambda$.\\
\noindent
For $-\varepsilon < r < \varepsilon$:\\
In this region, we have $\mathcal{L}_{HE}(r) = 0$, which is trivially bounded by $\lambda$.\\
\noindent
For $r \geq \varepsilon$:\\
Similar to the first region, we can easily demonstrate that in this region $\mathcal{L}_{HE}(r)$ is also bounded by $\lambda$.\\
Hence, in all three regions, we have shown that $\mathcal{L}_{HE}(r)$ is bounded by $\lambda$ for the specified parameter values $a > 0$, $\lambda > 0$, and $\varepsilon > 0$.
\end{proof}
\noindent
\textbf{Property 6}:
$\mathcal{L}_{HE}(\cdot)$ is smooth, indicating that it is differentiable and suitable for gradient-based optimization algorithms.
\begin{proof}
 The loss function $\mathcal{L}_{HE}(\cdot)$ is meticulously constructed by combining continuous functions and exponential terms, ensuring its overall differentiability. Nevertheless, to dispel any concerns regarding its differentiability specifically at $r = \varepsilon$ and $r = -\varepsilon$, we provide a thorough demonstration limited to these particular points.\\
At $r=\varepsilon:$
\begin{align*}
\text{L.H.D.}&= \lim _{h \rightarrow 0} \frac{\mathcal{L}_{HE}(\varepsilon - h) - \mathcal{L}_{HE}(\varepsilon)  }{-h} \nonumber\\
&= \lim _{h \rightarrow 0} \frac{0 - 0}{-h} \hspace{6.5cm}(\text{Using equation}~ (\ref{Proposed Loss}))\nonumber \\
&= 0 \nonumber.
\end{align*}
\begin{align*}
\text{R.H.D.}&= \lim _{h \rightarrow 0} \frac{\mathcal{L}_{HE}(\varepsilon + h) - \mathcal{L}_{HE}(\varepsilon)  }{h} \nonumber\\
&= \lim _{h \rightarrow 0} \frac{ \lambda \left[  1-\{a(\varepsilon + h-\varepsilon)+1\} e^{-a(\varepsilon + h-\varepsilon)}   \right] - 0}{h} \hspace{0.4cm}(\text{Using equation}~ (\ref{Proposed Loss}))\nonumber \\
&= \lim _{h \rightarrow 0} \frac{ \lambda \left[  1-\{ah+1\} e^{-ah}   \right] }{h} \hspace{6cm} (\frac{0}{0}~\text{form}) \nonumber\\
&= \lim _{h \rightarrow 0} -\lambda \left[-a(ah+1)e^{-ah}+ae^{-ah}\right] \hspace{2cm} (\text{Using L'hospital rule}) \nonumber\\
&= -\lambda \left[ -a + a \right] \nonumber\\
&= 0. \nonumber
\end {align*}
Thus, at $r=\varepsilon$, the left-hand derivative and right-hand derivative are both equal. Hence, $\mathcal{L}_{HE}(\cdot)$ is differentiable at $r=\varepsilon$.\\
At $r=-\varepsilon:$
\begin{align*}
\text{L.H.D.}&= \lim _{h \rightarrow 0} \frac{\mathcal{L}_{HE}(-\varepsilon - h) - \mathcal{L}_{HE}(-\varepsilon)  }{-h} \nonumber\\
&= \lim _{h \rightarrow 0} \frac{\lambda \left[  1-\{-a(-\varepsilon - h+\varepsilon)+1\} e^{a(-\varepsilon - h+\varepsilon)}   \right] - 0}{-h} \hspace{0.4cm}(\text{Using equation}~ (\ref{Proposed Loss}))\nonumber \\
&= \lim _{h \rightarrow 0} \frac{\lambda \left[  1-\{-a(- h)+1\} e^{a(- h)}   \right]}{-h} \hspace{5.3cm} (\frac{0}{0}~\text{form}) \nonumber\\
&= \lim _{h \rightarrow 0} \lambda \left[-a(ah+1)e^{-ah}+ae^{-ah}\right] \hspace{2.8cm} (\text{Using L'hospital rule}) \nonumber\\
&= \lambda \left[ -a + a \right] \nonumber\\
&= 0. \nonumber
\end{align*}
\begin{align*}
\text{R.H.D.}&= \lim _{h \rightarrow 0} \frac{\mathcal{L}_{HE}(-\varepsilon + h) - \mathcal{L}_{HE}(-\varepsilon)  }{h} \nonumber\\
&= \lim _{h \rightarrow 0} \frac{ 0 - 0}{h} \hspace{7.2cm}(\text{Using equation}~ (\ref{Proposed Loss}))\nonumber \\
&= 0. \nonumber
\end{align*}
Thus, at $r=-\varepsilon$, the left-hand derivative and right-hand derivative are both equal. Hence, $\mathcal{L}_{HE}(\cdot)$ is differentiable at $r=-\varepsilon$. Consequently, we have established that $\mathcal{L}_{HE}(\cdot)$ is differentiable throughout its domain.
\end{proof}

To enable gradient-based optimization, we can derive the derivative of $\mathcal{L}_{HE}(\cdot)$ with respect to $r$:
\begin{align}\label{Derivate ofProposed Loss}
\frac{\partial\mathcal{L}_{HE} (r, a, \lambda)}{\partial r}=
\begin{cases}
\lambda a^2 (r+\varepsilon) e^{a(r+\varepsilon)}, & r < -\varepsilon, \\
0, & -\varepsilon \leq r \leq \varepsilon,\\
\lambda a^2 (r-\varepsilon) e^{-a(r-\varepsilon)}, & r > \varepsilon.
\end{cases}
\end{align}
The derivative of $\mathcal{L}_{HE}(\cdot)$ is visualized for different values of parameters in Figures \ref{fig:third} and \ref{fig:four}. The shape of the derivative provides valuable insights into our loss function and its behavior during minimization using gradient descent or some related method. For all values of $a$ and $\lambda$, the derivative is $0$ when $|r| \leq \varepsilon$, indicating no sensitivity to minor residuals. However, as the magnitude of the residual exceeds $\varepsilon$, the derivative increases with the growth of $a$ and $\lambda$. Further, as the magnitude of the residual progressively increases beyond a certain threshold, the magnitude of the derivative begins to decrease, ultimately converging towards zero. Consequently, as the magnitude of an outlier's residual grows, its influence on the gradient descent process diminishes, and once it surpasses a certain threshold, the outlier's impact becomes nearly negligible.

\subsection{Comparative Analysis of HawkEye Loss Function with baseline loss functions:}
\begin{figure*}
\centering
    \subcaptionbox{     \label{fig:Hawk_vs_least}} { %
    \includegraphics[width=0.48\textwidth,keepaspectratio]{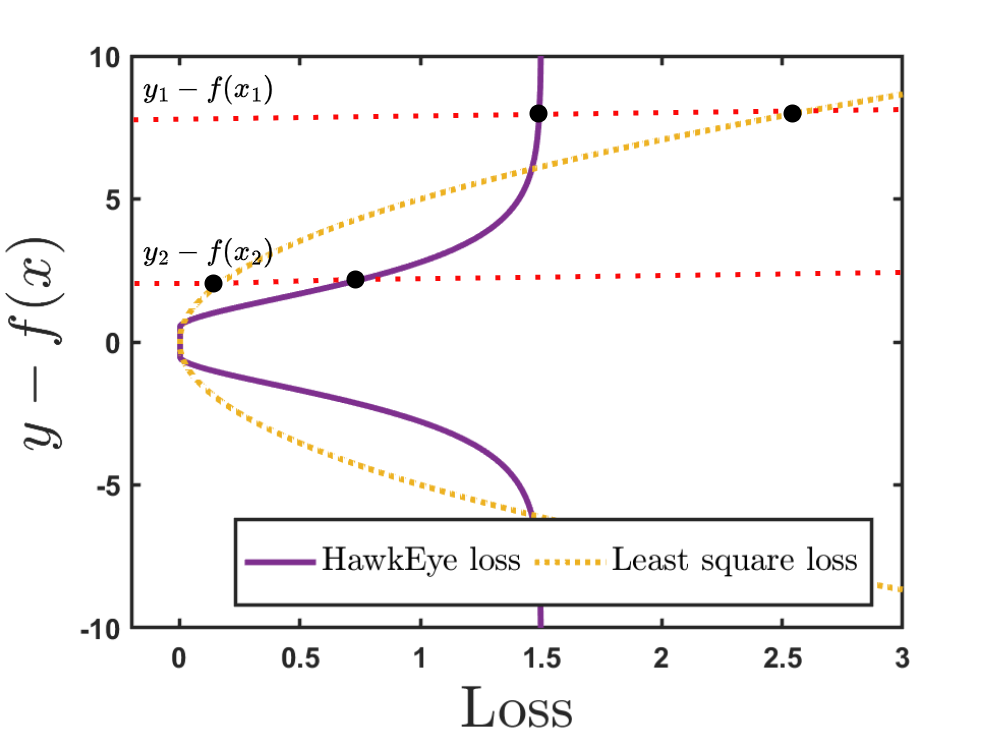}}
      \hfill
      \subcaptionbox{   \label{fig:Hawk_vs_insensitive}} { %
      \includegraphics[width=0.48\textwidth,keepaspectratio]{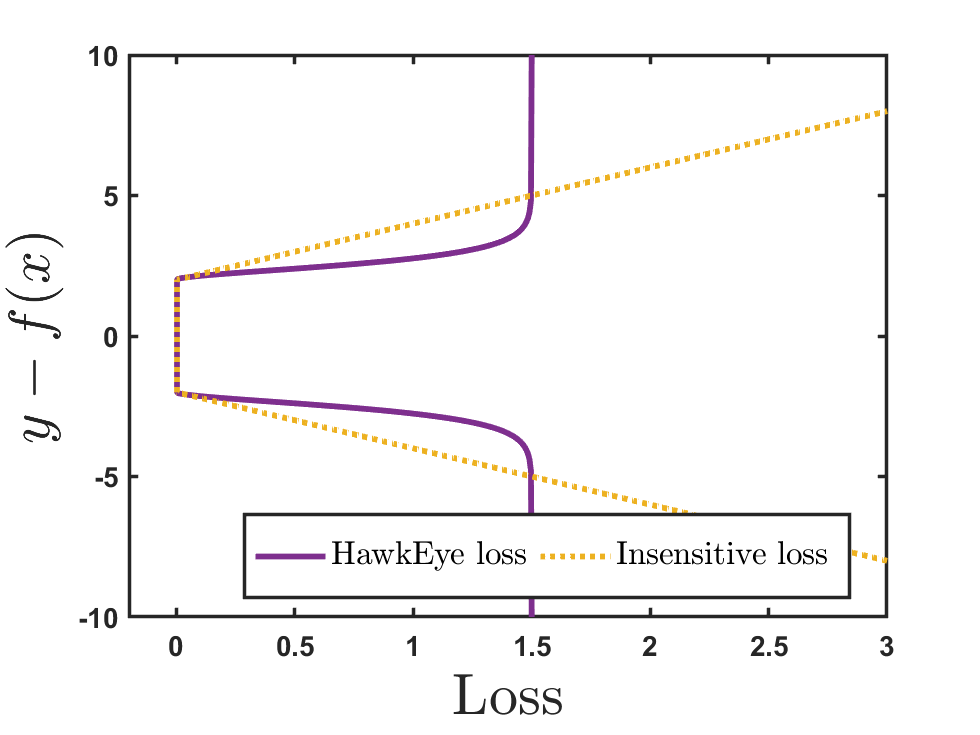}}
\\
      \subcaptionbox{  \label{fig:Hawk_vs_canal}} { %
      \includegraphics[width=0.48\textwidth,keepaspectratio]{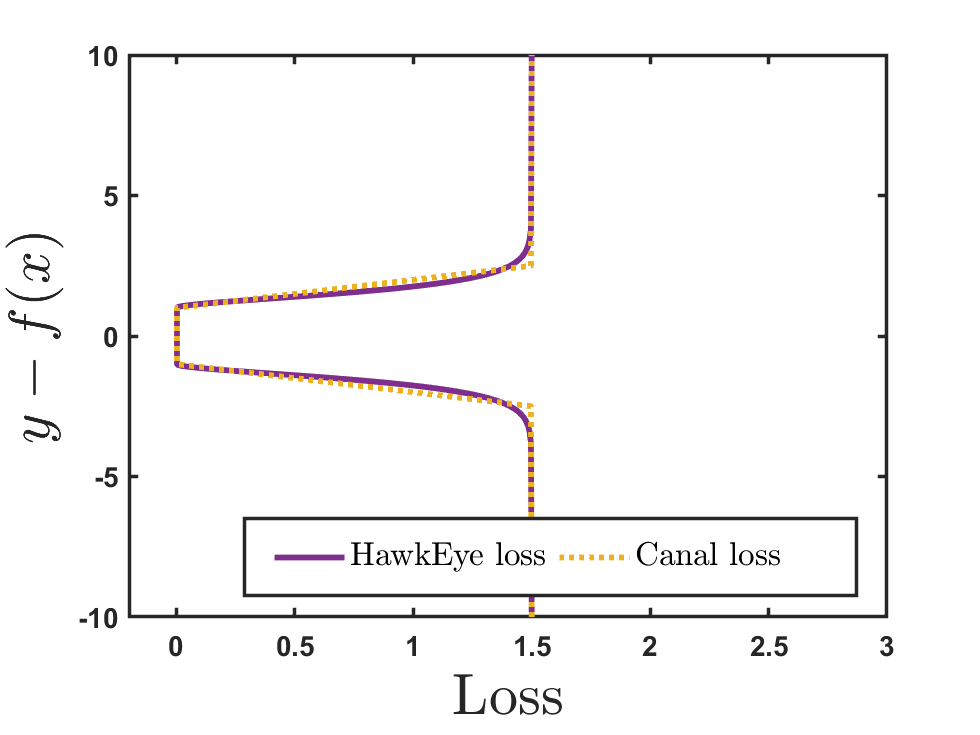}}
      \hfill
      \subcaptionbox{  \label{fig:Hawk_vs_boundedleast}} { %
    \includegraphics[width=0.48\textwidth,keepaspectratio]{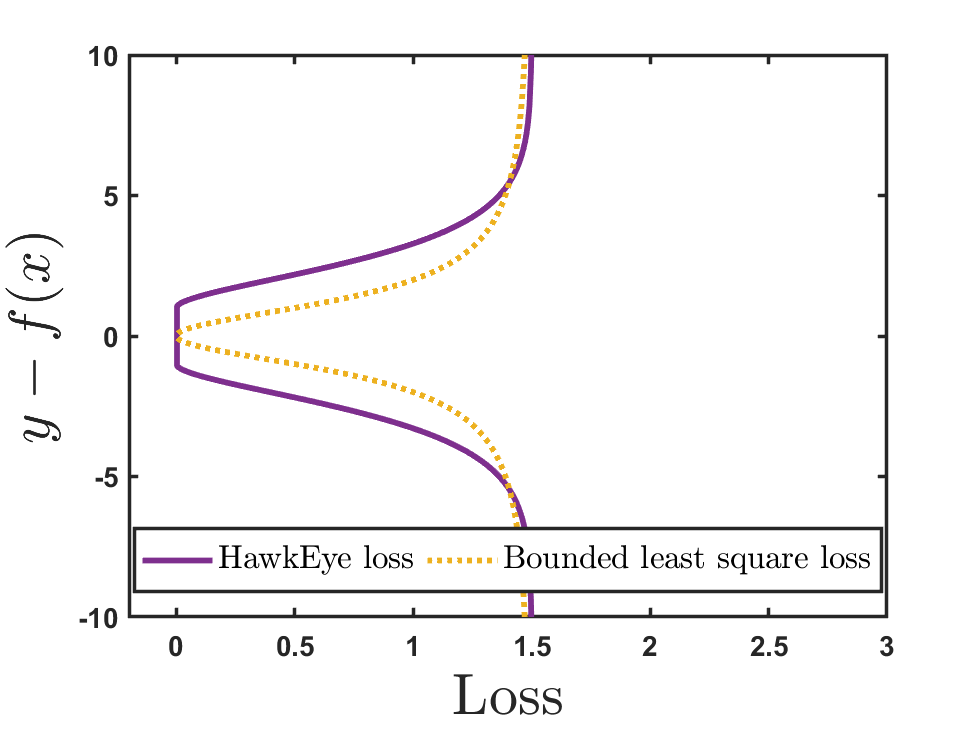}}
      \caption{Visual comparison of HawkEye loss and baseline loss functions. Subfigures (a), (b), (c), and (d) demonstrate the comparison of HawkEye loss function with the least square loss, insensitive loss, canal loss, and bounded least square loss, respectively.}
    \label{fig:HawkEye_vs_baseline}
 \end{figure*}

In this subsection, we conduct a comparative analysis of the proposed HawkEye loss function with some established baseline loss functions. Each baseline loss function is examined individually to highlight its characteristics and performance relative to the HawkEye loss function. For the sake of clarity, we presented visual depictions in Figure \ref{fig:HawkEye_vs_baseline}. 

\begin{enumerate}
    \item \textbf{HawkEye loss} ($\mathcal{L}_{HE}$) $\emph{vs}$ \textbf{Least square loss} ($\mathcal{L}_{ls}$): Consider a data point $(x_1, y_1)$ with a noisy label where the target value $y_1$ is mislabeled for some reason.  As a consequence, the target value $y_1$ deviates significantly from the actual model prediction $f(x_1)$, resulting in a large absolute difference of $|y_1 - f(x_1)|$. In such cases, the least squares loss becomes exceedingly large, and minimizing this loss would have a notorious repercussion on the overall performance of the model. In contrast, when we apply the HawkEye loss to such a data point in similar circumstances, it tends to approach a constrained value, and we note that $\mathcal{L}_{HE}(x_1)$ is notably less than $\mathcal{L}_{ls}(x_1)$ (see Figure \ref{fig:Hawk_vs_least}). This observation underscores that the HawkEye loss function effectively mitigates the influence of noisy data points, highlighting its robustness in the presence of such outliers. Now, let's turn our attention to another data point $(x_2, y_2)$ that is clean and exhibits a small absolute difference $|y_2 - f(x_2)|$. Such points are inherently more amenable to misestimation and thus warrant closer attention. Notably, when we examine the least squares loss for this point, it is less than the corresponding HawkEye loss, as evidenced by $\mathcal{L}_{HE}(x_2)$ $>$ $\mathcal{L}_{ls}(x_2)$. Thus, it can be confidently asserted that, with appropriately selected parameters, the HawkEye loss function not only demonstrates resilience in the presence of noisy data points but also exhibits a capacity for affording greater attention to clean, non-outlier data points.
    \item \textbf{HawkEye loss} ($\mathcal{L}_{HE}$) $\emph{vs}$ \textbf{Insensitive loss} ($\mathcal{L}_{\varepsilon}$): Similar to the least squares loss function, the insensitive loss function imposes a substantial penalty on samples exhibiting large deviations, making it less robust in the presence of outliers when compared to the HawkEye loss. Furthermore, for a judiciously chosen parameter value, the HawkEye loss exhibits a propensity to allocate greater emphasis to non-outlier points compared to the insensitive loss function. It is also important to highlight that the HawkEye loss function surpasses the insensitive loss function in terms of smoothness. The inherent smoothness property of the HawkEye loss is an added advantage in optimization processes.
    \item \textbf{HawkEye loss} ($\mathcal{L}_{HE}$) $\emph{vs}$ \textbf{Canal loss} ($\mathcal{L}_{canal}$): The Canal loss is characterized by its capability to constrain the loss to a predefined value through the selection of parameters $\theta$ and $\varepsilon$. Similarly, the HawkEye loss function shares this property, as it determines an upper bound $\lambda$ for penalty and restricts the loss to stop raising after a certain deviation. However, in the context of non-outlier data points, the canal loss maintains a fixed shape. In contrast, the proposed HawkEye loss function exhibits greater versatility, as it can adapt its shape to suit diverse situations by adjusting the shape parameter value $a$. It's also essential to underscore that the HawkEye loss outshines the canal loss not only in adaptability but also in terms of smoothness. Hence the HawkEye loss, with its smooth profile, presents a more advantageous characteristic compared to the canal loss.
    \item \textbf{HawkEye loss} ($\mathcal{L}_{HE}$) $\emph{vs}$ \textbf{Bounded least square loss} ($\mathcal{L}_{bls}$): The bounded least square loss function is equipped with the capability to confine the loss within a predefined threshold by adjusting the parameter $t$. Simultaneously, it offers the flexibility to adapt its shape for non-outlier data points through the modulation of the parameter $\theta$. Intriguingly, the HawkEye loss function also shares both of these remarkable characteristics. A notable highlight is that the HawkEye loss introduces an additional dimension by incorporating the concept of an insensitive zone. An enlightening point of distinction lies in the fact that the HawkEye loss function is the first loss function in SVR literature that not only possesses the attributes of smoothness and boundedness but also concurrently exhibits an insensitive zone.
\end{enumerate}

 We have also summarized the key attributes of the existing baseline loss functions and the proposed HawkEye loss function in Table \ref{tab:comparison_of_loss_function}.

\begin{table}[]
\caption{Demonstrate the characteristics of baseline loss functions used in regression and the proposed HawkEye loss.}
\label{tab:comparison_of_loss_function}
\resizebox{\textwidth}{!}{%
\begin{tabular}{|l|l|l|l|l|l|}
\hline
\textbf{Loss function $\downarrow$\textbackslash{}Characteristics $\rightarrow$} & \textbf{Robust} & \textbf{Insensitive zone} & \textbf{Bounded} & \textbf{Convex} & \textbf{Smooth} \\ \hline
\textbf{Least square loss}                     &~~~~{\color{red}\ding{55}}  &~~~~~~~~~~~{\color{red}\ding{55}}  &~~~~~{\color{red}\ding{55}}  &~~~~{\color{blue}\ding{51}}  & ~~~~{\color{blue}\ding{51}} \\ \hline
\textbf{Absolute loss}                         &~~~~{\color{red}\ding{55}}  &~~~~~~~~~~~{\color{red}\ding{55}}  &~~~~~{\color{red}\ding{55}}  & ~~~~{\color{blue}\ding{51}} &~~~~{\color{red}\ding{55}}  \\ \hline
\textbf{Huber loss}                            &~~~~{\color{red}\ding{55}}  &~~~~~~~~~~~{\color{red}\ding{55}}  & ~~~~~{\color{red}\ding{55}} &~~~~{\color{blue}\ding{51}}  &~~~~{\color{blue}\ding{51}}  \\ \hline
\textbf{Insensitive loss}                      &~~~~{\color{red}\ding{55}}  & ~~~~~~~~~~~{\color{blue}\ding{51}} &~~~~~{\color{red}\ding{55}}  & ~~~~{\color{blue}\ding{51}} & ~~~~{\color{red}\ding{55}} \\ \hline
\textbf{Ramp insensitive loss}                 &~~~~{\color{blue}\ding{51}}  & ~~~~~~~~~~~{\color{blue}\ding{51}} &~~~~~{\color{blue}\ding{51}}  &~~~~{\color{red}\ding{55}}  &~~~~{\color{red}\ding{55}}  \\ \hline
\textbf{Non-convex least square loss}          &~~~~{\color{blue}\ding{51}}  &~~~~~~~~~~~{\color{red}\ding{55}}  &~~~~~{\color{blue}\ding{51}}  &~~~~{\color{red}\ding{55}}  & ~~~~{\color{red}\ding{55}}  \\ \hline
\textbf{Ramp insensitive least square loss}    &~~~~{\color{blue}\ding{51}}  &~~~~~~~~~~~{\color{blue}\ding{51}}  &~~~~~{\color{blue}\ding{51}}  &~~~~{\color{red}\ding{55}}  &~~~~{\color{red}\ding{55}}  \\ \hline
\textbf{Quadratic non-convex insensitive loss} &~~~~{\color{blue}\ding{51}}  &~~~~~~~~~~~{\color{blue}\ding{51}}  &~~~~~{\color{blue}\ding{51}}  &~~~~{\color{red}\ding{55}}  & ~~~~{\color{red}\ding{55}}  \\ \hline
\textbf{Canal loss}                            &~~~~{\color{blue}\ding{51}}  &~~~~~~~~~~~{\color{blue}\ding{51}}  &~~~~~{\color{blue}\ding{51}}  &~~~~{\color{red}\ding{55}}  & ~~~~{\color{red}\ding{55}} \\ \hline
\textbf{Bounded least square loss}                   &~~~~{\color{blue}\ding{51}}  &~~~~~~~~~~~{\color{red}\ding{55}}  &~~~~~{\color{blue}\ding{51}}  & ~~~~{\color{red}\ding{55}} & ~~~~{\color{blue}\ding{51}} \\ \hline
\textbf{HawkEye loss (Proposed)}                  &~~~~{\color{blue}\ding{51}}  &~~~~~~~~~~~{\color{blue}\ding{51}}  &~~~~~{\color{blue}\ding{51}}  &~~~~{\color{red}\ding{55}}  &~~~~{\color{blue}\ding{51}}  \\ \hline
\end{tabular}%
}
\end{table}

\subsection{SVR with HawkEye loss function}
In this subsection, we incorporate the newly proposed HawkEye loss function into the least squares framework of SVR and establish a novel support vector regressor that manifests robustness against outliers and desirable smoothness characteristics. We refer to this advanced regressor as the HE-LSSVR. For the sake of simplicity and without loss of generality, we use the terminology $w$ for $\left[w^\intercal,b\right]$ and $x_i$ for $\left[x_i,1\right]^\intercal$ throughout the paper. The formulation of the proposed HE-LSSVR is given as:
\begin{align} \label{proposedSVR}
\underset{ w, \xi}{min}  \hspace{0.5cm}~& \frac{1}{2}\|w\|^2+C\sum_{i=1}^N \mathcal{L}_{HE}(\xi_i), \nonumber \\
\text { subject to }\hspace{0.2cm}  & \xi_i = y_i - \left(w^\intercal \phi(x_i)\right), ~\forall~ i=1,2, \ldots,N,
\end{align}
where $\frac{1}{2}\|w\|^2$ is the regularizer term, $\sum_{i=1}^N \mathcal{L}_{HE}(\xi_i)$ is the empirical risk corresponds to the proposed HawkEye loss function, $C>0$ is a regularization parameter, $\xi_i$ indicates the regression error of $i^{th}$ input, and $\phi(\cdot)$ represents the high-dimensional feature mapping associated with the kernel function. Let us assume that $\mathcal{K} : \mathbb{R}^m \times \mathbb{R}^m \rightarrow \mathbb{R}$ is a Mercer kernel and $H_\mathcal{K}$ is a reproducing kernel Hilbert space induced by $\mathcal{K}$. In practice, kernel functions are often employed to address nonlinear problems by formulating the relevant dual problems. However, due to the non-convex nature of equation (\ref{proposedSVR}), this approach can be quite formidable in our scenario. In this case, to enhance the capacity of HE-LSSVR for nonlinear adaptation, we use the representer theorem \cite{dinuzzo2012representer}, which allows us to express $w$ in equation (\ref{proposedSVR}) as follows:
\begin{align} \label{representer theorem}
    w= \sum_{k=1}^N \alpha_k \phi(x_k),
\end{align}
where $\alpha = \left(\alpha_1, \ldots, \alpha_N \right)^\intercal$ is the coeffecient vector. We note that the minimization problem (\ref{proposedSVR}) works in a reproducing kernel Hilbert space, and the penalty term is non-decreasing, so one can apply the representer theorem to the optimization problem (\ref{proposedSVR}).\\
Substituting the value of $w$ from (\ref{representer theorem}) into (\ref{proposedSVR}), the primal optimization problem (\ref{proposedSVR}) can be  transformed into the following unconstrained optimization problem:
\begin{align} \label{Proposed_Problem}
\underset{\alpha}{min}~ \sum_{i=1}^N \sum_{k=1}^N \frac{1}{2} \alpha_i \alpha_k \mathcal{K}\left(x_i, x_k\right)+C\sum_{i=1}^N \mathcal{L}_{HE}(\xi_i),
\end{align}
where $\xi_i = y_i - \sum_{k=1}^N \alpha_k \mathcal{K}\left(x_i, x_k\right)$ and $\mathcal{K}\left(x_i, x_k\right)= \phi\left(x_{i}\right) \cdot \phi\left(x_{k}\right)$ is the kernel function.\\

\subsection{Adaptive moment estimation (Adam) algorithm for HE-LSSVR}
The smoothness characteristic of the non-convex HE-LSSVR model enables the utilization of gradient-based algorithm to optimize the model (\ref{Proposed_Problem}). Leveraging gradient-based optimization offers several advantages, notably expediting convergence during the training process. This is particularly valuable since gradient-based algorithms typically exhibit faster convergence compared to quadratic programming solvers \cite{bottou2018optimization}. In this paper, we employ the adaptive moment estimation (Adam) \cite{kingma2014adam} optimization technique  to address the HE-LSSVR problem (\ref{Proposed_Problem}). It effectively combines the strengths of two popular optimization algorithms, AdaGrad \cite{duchi2011adaptive} and RMSProp \cite{tieleman2012lecture}. There are mainly three reasons to adopt Adam algorithm. Firstly, Adam employs adaptive learning rates for each weight parameter. This adaptability ensures that the learning rate is neither too high, which could lead to convergence issues, nor too low, which could slow down convergence. Thus contributing to faster and more stable convergence. Secondly, Adam incorporates the concept of momentum, which improves convergence in complex optimization landscapes. By maintaining a moving average of past gradients, it navigates these landscapes more efficiently and accelerates the convergence process.
Finally, Adam utilizes exponentially decreasing averages for previous gradients and squared gradients, providing flexibility to adapt to changing gradient behavior. This feature is well-suited for non-convex optimization tasks, where gradient characteristics may vary significantly throughout the training process.\\
Now, to begin with, let the objective function of (\ref{Proposed_Problem}) be $H(\alpha)$. Then, the procedure for applying the Adam algorithm is as follows:
\begin{enumerate}
\item \textbf{Initialization:} Start by initializing the algorithm with appropriate values for hyperparameters:\\
Learning Rate ($\gamma$): This is the step size for updating model parameters.
$\beta_1$ and $\beta_2$: These are constants for computing moving averages of gradients and squared gradients.
$\delta$: A small constant added to prevent division by zero.
\item \textbf{Initialize Moments:} Initialize the values of the first moment estimate ($m$) and the second moment estimate ($v$). These moments are used to keep track of past gradients and squared gradients, respectively.
\item \textbf{Iterative Optimization:} At each iteration $t$:

\begin{itemize}
    \item Compute the gradient of the objective function $H(\alpha)$ with respect to the model parameters $\alpha$.
\begin{align} \label{ non-linear gradient}
\nabla H(\alpha_t)=\mathcal{K} \alpha +C \sum_{i=1}^N \frac{\partial\mathcal{L}_{HE}}{\partial \alpha},
\end{align}
where 
\begin{align*}
\frac{\partial\mathcal{L}_{HE}}{\partial \alpha}=
\begin{cases}
-\lambda a^2 (\xi_i+\varepsilon)  e^{a(\xi_i+\varepsilon)} \mathcal{K}_i, & \xi_i < -\varepsilon, \\
0, & -\varepsilon \leq \xi_i \leq \varepsilon,\\
-\lambda a^2 (\xi_i-\varepsilon) e^{-a(\xi_i-\varepsilon)} \mathcal{K}_i, & \xi_i > \varepsilon.
\end{cases}
\end{align*}
Here $\mathcal{K}_i$ represent the $i^{th}$ row of the kernel matrix $\mathcal{K}$.
\item Update the first moment estimate:
\begin{align} \label{first moment estimate}
m_t= \beta_1  m_{t-1} + \left(1-\beta_1\right) \nabla H(\alpha_t).
\end{align}
\item Update the second moment estimate:
\begin{align} \label{second moment estimate}
v_t= \beta_2 v_{t-1} + \left(1-\beta_2\right) \nabla H(\alpha_t)^2.
\end{align}
\item Correct the bias in moments (due to initialization):
\begin{align} \label{corrected first moment}
\hat{m_t}= \frac{m_t}{\left(1-\beta_1^t\right)},
\end{align}
\begin{align} \label{corrected second moment}
\hat{v_t}= \frac{v_t}{\left(1-\beta_2^t\right)}.
\end{align}
\item Update model parameter:
\begin{align} \label{update rule}
\alpha_t= \alpha_{t-1} - \gamma \frac{\hat{m_t}}{\sqrt{\hat{v_t}+\delta}}.
\end{align}
\end{itemize}
\item \textbf{Convergence Criterion:} Monitor the convergence of the optimization process. A common stopping criterion is a maximum number of iterations or a target value for the objective function $H(\alpha)$.
\item \textbf{Termination:} When the algorithm meets the convergence criterion, terminate the optimization process, and the final model parameters $\alpha$ represent the optimal solution for the problem $H(\alpha)$.
\end{enumerate}
The structure of the Adam algorithm to solve the HE-LSSVR model is clearly described in Algorithm \ref{algorithm}. Once the optimal $\alpha$ is obtained, the following decision function can be utilized to predict the target value of a new sample $x$.
\begin{align} \label{decision function}
f(x)= \sum_{k=1}^N \alpha_{k}~\mathcal{K}\left(x, x_k\right).
\end{align}

\begin{figure}
\centering
    { %
\includegraphics[width=1.3\textwidth,keepaspectratio]{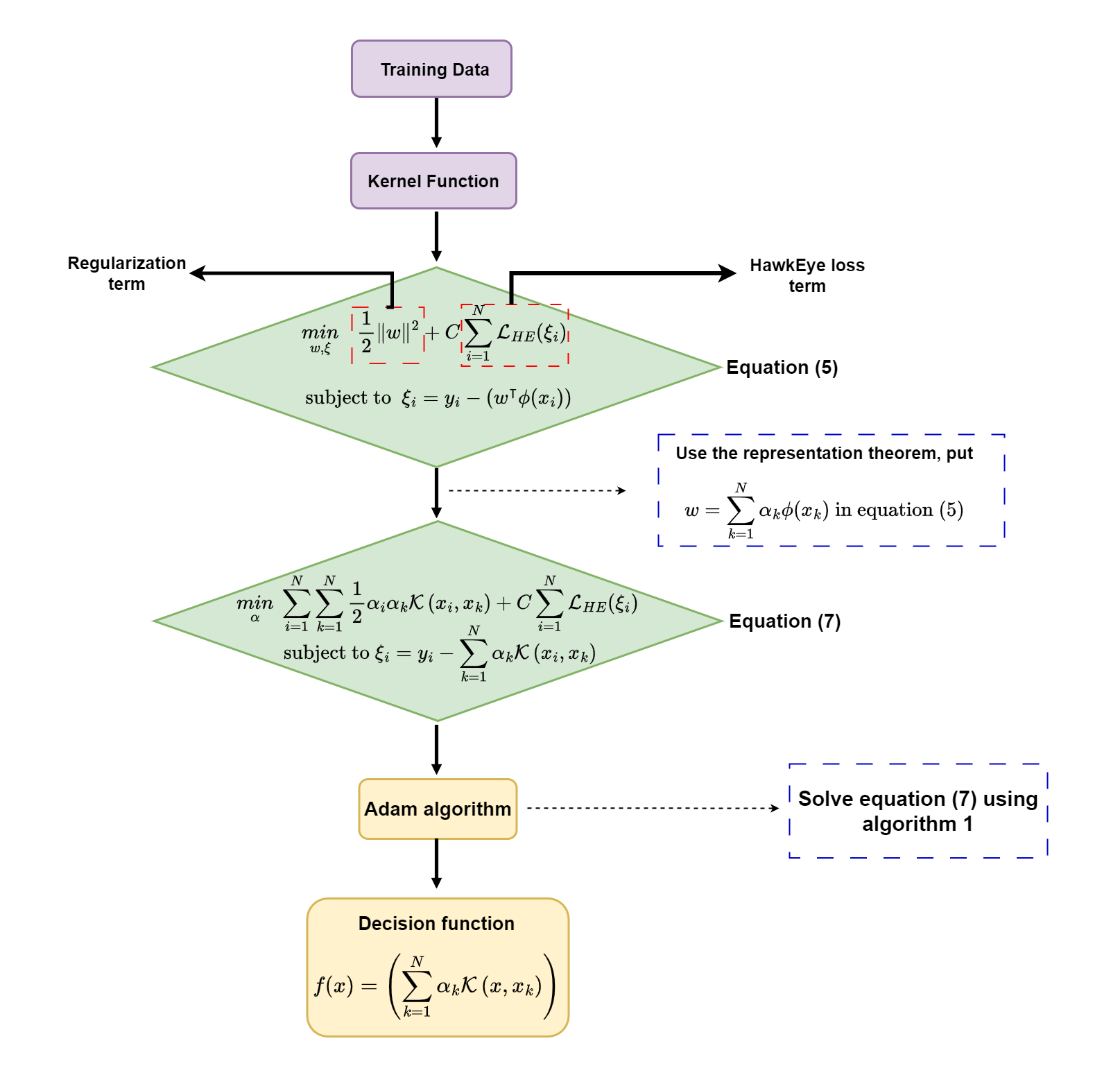}
    }
    \caption{Flowchart illustrating the key stages of the HE-LSSVR model, showcasing the transformation from original input data to the decision function, highlighting the strategic integration of the HawkEye loss function and the application of representer theorem and Adam algorithm.}
    \label{flowchart}
\end{figure}

For delineating the intricacies of the proposed HE-LSSVR model, in Figure \ref{flowchart}, we present a meticulously crafted flowchart that captures the essence of our methodology. This visual narrative serves not only as a roadmap for our methodology but also as a testament to the strategic interplay of components that positions HE-LSSVR as a cutting-edge regression model.


\begin{tiny}
\begin{algorithm}
  \caption{Adam algorithm for HE-LSSVR}
  \label{algorithm}
   \begin{algorithmic}
  \State \textbf{Input:}
     \State The dataset:  $\left\{x_i,y_i\right\}_{i=1}^N$, $y_i \in \mathbb{R}$;
     \State The parameters: Regularization parameter $C$, HawkEye loss parameters $\lambda$, $a$, and $\varepsilon$, mini-batch size $s$, decay rates $\beta_1$ and $\beta_2$, learning rate $\gamma$, constant $\delta$, maximum iteration number $T$;    
      \State Initialize: $\alpha_0$ and $t$;
     \State \textbf{Output:}
       \State The classifiers parameters: $\alpha$;
       \State $1:$ Select $s$ samples $\left\{x_i,y_i\right\}_{i=1}^s$ uniformly at random.
\State $2:$  Computing $\xi_i$ :
\begin{align}
\xi_i = y_i - \sum_{k=1}^s \alpha_k \mathcal{K}\left(x_i, x_k\right), \quad i=1, \ldots, s;
\end{align}
\State $3:$ Compute $\nabla H(\alpha_t)$: (\ref{ non-linear gradient});
\State $4:$ Compute $m_t$: (\ref{first moment estimate});
\State $5:$ Compute $v_t$: (\ref{second moment estimate});
\State $6:$ Compute $\hat{m_t}$: (\ref{corrected first moment});
\State $7:$ Compute $\hat{v_t}$: (\ref{corrected second moment});
\State $8:$ Updating solution $\alpha_t$: (\ref{update rule});
\State $9:$ Updating current iteration number: $t=t+1$.
\State \textbf{Until:}
\State $t=T$
\State \textbf{Return:} $\alpha_t$.
 \end{algorithmic}
\end{algorithm}
\end{tiny}

\section{Numerical Experiments and Result Analysis}
In this section, we conduct extensive experiments on UCI, syntetic and time series datasets to compare the proposed HE-LSSVR model with existing state-of-the-art models, which include support vector regression (SVR) \cite{drucker1996support}, least squares support vector regression (LS-SVR) \cite{suykens1999least}, and bounded least squares support vector regression (BLSSVR) \cite{fu2023robust}.
\subsection{Experimental Setup and Parameter Setting}
All the experiments are executed using MATLAB R2023a on Window 10 running on a PC with configuration Intel(R) Xeon(R) Gold 6226R CPU @ 2.90GHz, 2901 Mhz, 16 Core(s), 32 Logical Processor(s) with 128 GB of RAM. The kernel function plays a pivotal role in addressing non-linear models. We use the radial basis function (RBF) kernel as it is widely recognized for its high performance in regression tasks \cite{chang2011libsvm}. The RBF kernel is defined as $\mathcal{K}\left(x_i,x_k\right)=\exp\left(-\left\|x_i-x_k\right\|^2 / \sigma^2\right)$, where $\sigma$ is an adjustable parameter. For each model, the regularization parameter $C$ and kernel parameter $\sigma$ are selected from the set $\{10^i\,|\,i=-6, -4,\ldots, 4, 6\}$. For SVR and the proposed HE-LSSVR, the insensitive zone parameter $\varepsilon$ is selected from the set $\{0.001, 0.005, 0.01, 0.05, 0.1, 0.15, 0.2, 0.25\}$. The loss hyperparameters for BLSSVR are adopted the same as in \cite{fu2023robust}. For the proposed HE-LSSVR, loss hyperparameters $\lambda$ and $a$ are selected from the ranges $\left[0.1:0.2:2\right]$ and  $\left[0.1:0.2:5\right]$, respectively.  The Adam algorithm parameters are empirically configured as follows: (\RNum{1}) initial model parameter $\alpha_{0}= 0.01$, (\RNum{2}) initial first moment $m_{0}=0.01$, (\RNum{3}) initial second moment $v_{0}=0.01$, (\RNum{4}) initial learning rate $\gamma$ is chosen from the set $\{0.0001,0.001, 0.01\}$, (\RNum{5}) exponential decay rate of first order $\beta_{1}=0.9$, (\RNum{6}) exponential decay rate of second order $\beta_{2}=0.999$, (\RNum{7}) division constant $\delta=10^{-8}$, (\RNum{8}) mini-batch size $s=2^{5}$, (\RNum{10}) maximum iteration number $T=1000$.\\
The performance of machine learning models is markedly affected by the choice of hyperparameters \cite{shawe2004kernel}. To optimize these hyperparameters effectively, we performed a tuning process using $k$-fold ($k = 5$) cross-validation and grid search. This procedure involved dividing the dataset into five distinct and non-overlapping subsets, commonly referred to as ``folds". During each iteration, one of these subsets is designated for testing purposes, while the remaining four are employed for training. For each distinct set of hyperparameters, we computed the root mean squared error (RMSE) separately on each fold. We then determined the best result for each set of hyperparameters.  Finally, one with the lowest RMSE is selected as the optimal parameter combination.

\subsection{Evaluation Metrics}
Let $y_i$, $f(x_i)$ represent the actual and predicted values of $x_i$, respectively. To assess the models' performance, we employ four evaluation metrics: root mean squared error (RMSE), mean absolute error (MAE), positive error ($\text{Error}_{pos}$), and negative error ($\text{Error}_{neg}$). The specific definitions of these metrics are listed in Table \ref{tab:metric-table}. We primarily use RMSE as the key comparative measure, while the other metrics serve as supplementary evaluation criteria. Further, to show the efficiency of the proposed BS-SSVR, we also performed a comparative analysis of the training time.

\begin{table}[]
\centering
\caption{The evaluation metrics.}
\label{tab:metric-table}
\begin{tabular}{ll}
\hline
Metrics  & Definitions \\ \hline
         &             \\
RMSE     & $\sqrt{\frac{1}{N} \sum_{i=1}^N\left(y_i-f\left(\mathbf{x}_i\right)\right)^2}$            \\
         &             \\
MAE      &  $\frac{1}{N} \sum_{i=1}^N\left|\left(y_i-f\left(\mathbf{x}_i\right)\right)\right| $           \\
         &             \\
$\text{Error}_{pos}$ &  $\frac{1}{N} \sum_{i=1, y_i \geq f\left(\mathbf{x}_i\right)}^N\left|y_i-f\left(\mathbf{x}_i\right)\right|$           \\
         &             \\
$\text{Error}_{neg}$ & $\frac{1}{N} \sum_{i=1, y_i<f\left(\mathbf{x}_i\right)}^N\left|y_i-f\left(\mathbf{x}_i\right)\right|$ \\
&     \\ \hline
\end{tabular}%
\end{table}

\subsection{Experiments on Benchmark UCI Datasets}
In this subsection, we use $18$ regression benchmark datasets from the UCI repository \cite{dua2017uci} to assess and compare the performance of the proposed HE-LSSVR against the baseline models. These datasets can be categorized into three groups: 1) 8 small-size datasets with samples ranging from 60 up to 500; 2) 4 medium-size datasets with samples ranging from 501 up to 1000; and 3) 6 large-size datasets with samples ranging from 1001 up to 40768. The detailed description of the datasets is listed in Table \ref{tab:dataset_description}. The experimental result of the proposed HE-LSSVR model and baseline models (SVR, LS-SVR, and BLSSVR) on UCI datasets in terms of metric RMSE, MAE, $\text{Error}_{pos}$, and $\text{Error}_{neg}$ is presented in Table \ref{tab:UCI-table}. The evaluation is primarily based on the RMSE, a critical metric where lower values indicate superior model performance. The proposed HE-LSSVR model demonstrates noteworthy performance, achieving the best RMSE values on $13$ out of $18$ datasets. Even on the remaining $5$ datasets, HE-LSSVR secures the second-best RMSE. This consistency across a diverse set of datasets underscores the prowess of the HE-LSSVR model. The average RMSE values further validate the effectiveness of HE-LSSVR. The existing SVR, LS-SVR, and BLSSVR models exhibit average RMSE values of $0.3113$, $0.6629$, and $0.3266$, respectively. In contrast, the proposed HE-LSSVR outperforms the baseline models with an average RMSE of $0.2736$. To ensure an accurate assessment of the models' performance, it is important to rank them individually for each dataset instead of relying solely on average RMSE. As exceptional performance on certain datasets may overshadow subpar performance on others. Table \ref{tab:Rank-table} presents the ranks assigned to each model based on their RMSE values across all datasets. The model with the lowest RMSE receives the lowest rank and vice-versa. The average rank for SVR, LSSVR, BLSSVR, and the proposed HE-LSSVR are $2.5294$, $3.8888$, $2$, and $1.2777$, respectively. It clearly demonstrates that the proposed HE-LSSVR model outperforms the baseline models, indicating its superior performance.
\par
To further support the effectiveness of the proposed HE-LSSVR model, we conducted a statistical analysis using the Friedman test, followed by the Nemenyi post hoc test. The Friedman test \cite{friedman1940comparison} is utilized to statistically assess the significance of various models. In this test, each individual model is ranked independently for each dataset. The null hypothesis posits that all models are essentially identical, implying that the average rank of each model is equivalent. The Friedman statistic adheres to the chi-squared $\chi^2_F$ distribution with degrees of freedom (d.f.) equal to $p-1$, where $p$ represents the number of models. This statistic is computed as follows:
\begin{align}{} \label{chisquareequation}
\chi_F^2=\frac{12 D}{p(p+1)}\left[\sum_e R_e^2-\frac{p(p+1)^2}{4}\right].
\end{align}
Here, $D$ signifies the total number of datasets, and $R_e$ denotes the mean rank of the $e^{th}$ model out of the $p$ models. However, the Friedman statistic is overly cautious in nature. To address this, \citet{iman1980approximations} introduced a more robust statistic:
\begin{align}{} \label{ffequation}
F_F=\frac{(D-1) \chi_F^2}{D(p-1)-\chi_F^2},
\end{align}
which follows $F$ distribution with $((p-1),(p-1)(D-1))$ d.f.. For $p=4$ and $D=18$, we obtain $\chi_F^2 = 23.2540$ and $F_F= 12.8575$. Also, $F_F$ is distributed with $(3,51)$ d.f.. From the statistical F-distribution table, at $5 \%$ level of significance, the value of $F_{(3,51)}= 2.68$. Since, $F_F > 2.68$, thus we reject the null hypothesis. Hence, substantial differences exist among the models. Next, we conduct the Nemenyi post hoc test \cite{demvsar2006statistical} to compare the models pairwise. The disparity in performance between the two models is considered significant if the corresponding average ranks display a noticeable difference surpassing a specific threshold value known as the critical difference ($C.D.$). When the discrepancy in mean ranks between the two models exceeds the $C.D.$, the model with the higher mean rank is deemed statistically superior to the model with the lower mean rank. The computation of $C.D.$ follows the following formula:
\begin{align}{}
    C.D.=q_\alpha \sqrt{\frac{p(p+1)}{6D}}.
\end{align}
Here, $q_\alpha$ is derived from the studentized range statistic divided by $\sqrt{2}$. This value is referred to as the critical value for the two-tailed Nemenyi test. From the distribution table, at $5\%$ level of significance, the value of $q_\alpha$ is $2.569$. Thus, after calculation, we obtain $C.D.= 1.1055$. Table \ref{tab:Nemenyi-table} presents the results of the Nemenyi post hoc test on UCI datasets. The average rank differences of the proposed HE-LSSVR from SVR, LS-SVR, and BLSSVR are $1.2517$, $2.6103$, and $0.7223$, respectively. Notably, the disparities in the average ranks for SVR and LSSVR exceed the $C.D.$; thus, according to the Nemenyi post hoc test, the proposed HE-LSSVR is significantly better than SVR and LS-SVR. It is noteworthy that the proposed HE-LSSVR model does not show a statistical difference from the BLSSVR; however, the proposed HE-LSSVR model outperformed the BLSSVR on $16$ out of $18$ datasets (see Table \ref{tab:UCI-table}). Considering these results, we can conclude that the HE-LSSVR model exhibits superiority over the baseline models.
\par
To show the efficacy of the proposed HE-LSSVR model in terms of computation, we also evaluated the training time of the models. Table \ref{tab:Time-table} illustrates the training times for both baseline models and the proposed HE-LSSVR models across different datasets. The 2D\_Planes dataset was left out because it had a memory error during training for the SVR model. The average training times for SVR, LS-SVR, BLSSVR, and HE-LSSVR are $34.7984$s, $11.7617$s, $0.0039$s, and	$0.0049$s, respectively. These results manifest the exceptional efficiency of the proposed HE-LSSVR, with significantly lower training times compared to SVR and LS-SVR. However, the proposed HE-LSSVR takes a slightly longer time to train than the BLSSVR, although the difference in training time is minor. It is crucial to note that the comparative evaluation of RMSE and training time demonstrates the overall performance of the models. In this regard, the proposed HE-LSSVR model stands out as it strikes a balance between RMSE and training time, making it superior to the baseline models.

\begin{table*}[htp]
\centering
\caption{Description of benchmark UCI datasets.}
\label{tab:dataset_description}
\begin{tabular}{lrr}
\hline
Datasets & \multicolumn{1}{l}{No. of samples} & \multicolumn{1}{l}{No. of features} \\ \hline
Airfoil\_Self\_Noise & 1503 & 5 \\
Delta\_Ailerons & 7129 & 5 \\
Forest\_Fires & 517 & 12 \\
Yacht\_Hydrodynamics & 308 & 6 \\
2D\_Planes & 40768 & 10 \\
Ailerons & 13750 & 40 \\
Parkinsons\_Telemonitoring & 5875 & 21 \\
Pole\_Telecomm & 15000 & 48 \\
2014\_2015 CSM dataset & 185 & 11 \\
cpu\_pref & 209 & 9 \\
hungary chickenpox & 522 & 19 \\
istanbul stock exchange data & 536 & 8 \\
mcs\_ds\_edited\_iter\_shuffled & 107 & 5 \\
qsar\_aquatic toxicity & 546 & 8 \\
Daily\_Demand\_Forecasting\_Orders & 60 & 12 \\
bodyfat & 252 & 14 \\
machine & 209 & 9 \\
slump\_test & 103 & 10 \\ \hline
\end{tabular}%
\end{table*}


\begin{table}[htp]
\centering
\caption{RMSE, MAE, $\text{Error}_{pos}$, and $\text{Error}_{neg}$ values on the benchmark UCI datasets for the proposed HE-LSSVR model and the baseline models.}
\label{tab:UCI-table}
\resizebox{\textwidth}{!}{
\begin{tabular}{lccccc}
\hline
Dataset & Metric & SVR\cite{drucker1996support} & LS-SVR\cite{suykens1999least} & BLSSVR\cite{fu2023robust} & HE-LSSVR$^{\dagger}$ \\ \hline
{2014\_2015 CSM dataset} & RMSE & 0.2604 & 0.3564 & 0.2547 & \textbf{0.2418} \\
 & MAE & 0.1916 & 0.3193 & 0.1674 & 0.1722 \\
 & $\text{Error}_{pos}$ & 0.2824 & 0.2588 & 0.2373 & 0.2626 \\
 & $\text{Error}_{neg}$ & 0.1625 & 0.3288 & 0.1295 & 0.134 \\ \hline
{cpu\_pref} & RMSE & \textbf{0.0332} & 0.3907 & 0.0379 & 0.0379 \\
 & MAE & 0.0231 & 0.3889 & 0.0298 & 0.0298 \\
 & $\text{Error}_{pos}$ & 0.0232 & - & 0.0238 & 0.0238 \\
 & $\text{Error}_{neg}$ & 0.023 & 0.3889 & 0.0395 & 0.0394 \\ \hline
{hungary chickenpox} & RMSE & 0.4892 & 0.7157 & \textbf{0.487} & \textbf{0.487} \\
 & MAE & 0.3607 & 0.6105 & 0.3484 & 0.3492 \\
 & $\text{Error}_{pos}$ & 0.4542 & 0.519 & 0.4442 & 0.4517 \\
 & $\text{Error}_{neg}$ & 0.3112 & 0.627 & 0.2861 & 0.2852 \\ \hline
{istanbul stock exchange data} & RMSE & 0.5783 & 0.5798 & 0.5709 & \textbf{0.5644} \\
 & MAE & 0.4502 & 0.4513 & 0.4443 & 0.4512 \\
 & $\text{Error}_{pos}$ & 0.4889 & 0.4835 & 0.4838 & 0.5079 \\
 & $\text{Error}_{neg}$ & 0.4054 & 0.4109 & 0.3967 & 0.3378 \\ \hline
{mcs\_ds\_edited\_iter\_shuffled} & RMSE & 0.1014 & 0.6715 & 0.0978 & \textbf{0.0812} \\
 & MAE & 0.0855 & 0.5937 & 0.0796 & 0.0623 \\
 & $\text{Error}_{pos}$ & 0.0781 & 0.0189 & 0.0878 & 0.0817 \\
 & $\text{Error}_{neg}$ & 0.0974 & 0.6224 & 0.0727 & 0.0342 \\ \hline
{qsar\_aquatic toxicity} & RMSE & 0.679 & 0.6224 & 0.5708 & \textbf{0.5348} \\
 & MAE & 0.5889 & 0.4765 & 0.5055 & 0.2557 \\
 & $\text{Error}_{pos}$ & 1.1452 & 1.0737 & 0.8531 & 0.2741 \\
 & $\text{Error}_{neg}$ & 0.5135 & 0.3737 & 0.45 & 0.0731 \\ \hline
{Daily\_Demand\_Forecasting\_Orders} & RMSE & 0.0426 & 0.4943 & 0.0493 & \textbf{0.0424} \\
 & MAE & 0.0361 & 0.4919 & 0.0437 & 0.0345 \\
 & $\text{Error}_{pos}$ & 0.0307 & - & 0.0535 & 0.0259 \\
 & $\text{Error}_{neg}$ & 0.047 & 0.4919 & 0.0367 & 0.0518 \\ \hline
{bodyfat} & RMSE & 0.0506 & 1.1625 & 0.0567 & \textbf{0.037} \\
 & MAE & 0.0381 & 1.0902 & 0.047 & 0.0282 \\
 & $\text{Error}_{pos}$ & 0.0559 & - & 0.0537 & 0.0265 \\
 & $\text{Error}_{neg}$ & 0.0284 & 1.0902 & 0.0422 & 0.0305 \\ \hline
{machine} & RMSE & \textbf{0.0329} & 0.3924 & 0.0382 & 0.0382 \\
 & MAE & 0.0223 & 0.3893 & 0.0299 & 0.0301 \\
 & $\text{Error}_{pos}$ & 0.0223 & - & 0.0242 & 0.0254 \\
 & $\text{Error}_{neg}$ & 0.0222 & 0.3893 & 0.0383 & 0.037 \\ \hline
{slump\_test} & RMSE & 0.0308 & 0.2741 & 0.0329 & \textbf{0.0251} \\
 & MAE & 0.0285 & 0.2261 & 0.0286 & 0.0209 \\
 & $\text{Error}_{pos}$ & 0.0309 & 0.1493 & 0.0321 & 0.0238 \\
 & $\text{Error}_{neg}$ & 0.027 & 0.2675 & 0.022 & 0.0197 \\ \hline
{Airfoil\_Self\_Noise} & RMSE & 0.2286 & 0.2967 & 0.2268 & \textbf{0.0864} \\
 & MAE & 0.1873 & 0.2702 & 0.1764 & 0.0505 \\
 & $\text{Error}_{pos}$ & 0.2617 & 0.2283 & 0.2531 & 0.0699 \\
 & $\text{Error}_{neg}$ & 0.1551 & 0.279 & 0.1334 & 0.0088 \\ \hline
{Delta\_Ailerons} & RMSE & 0.3653 & 0.507 & 0.3637 & \textbf{0.1881} \\
 & MAE & 0.3026 & 0.4186 & 0.3003 & 0.144 \\
 & $\text{Error}_{pos}$ & 0.2924 & 0.1732 & 0.3115 & 0.1378 \\
 & $\text{Error}_{neg}$ & 0.311 & 0.4846 & 0.2895 & 0.1525 \\ \hline

\end{tabular}}
\end{table}

\begin{table}[htp]
\ContinuedFloat
\centering
\caption{(Continued) RMSE, MAE, $\text{Error}_{pos}$, and $\text{Error}_{neg}$ values on the benchmark UCI datasets for the proposed HE-LSSVR model and the baseline models.}
\label{tab:UCI-table}
\resizebox{\textwidth}{!}{
\begin{tabular}{lccccc}
\hline
Dataset & Metric & SVR\cite{drucker1996support} & LS-SVR\cite{suykens1999least} & BLSSVR\cite{fu2023robust} & HE-LSSVR$^{\dagger}$ \\ \hline
{Forest\_Fires} & RMSE & 0.1487 & 0.4088 & \textbf{0.1183} & \textbf{0.1183} \\
 & MAE & 0.0994 & 0.394 & 0.0935 & 0.0936 \\
 & $\text{Error}_{pos}$ & 0.0322 & 0.6846 & 0.0653 & 0.0655 \\
 & $\text{Error}_{neg}$ & 0.1109 & 0.3883 & 0.1654 & 0.1653 \\ \hline
{Yacht\_Hydrodynamics} & RMSE & 1.0056 & 1.1753 & \textbf{0.5702} & 0.5998 \\
 & MAE & 0.9105 & 0.8238 & 0.532 & 0.4612 \\
 & $\text{Error}_{pos}$ & 1.0019 & 0.8958 & 0.5362 & 0.3786 \\
 & $\text{Error}_{neg}$ & 0.8248 & 0.2683 & 0.5309 & 0.5801 \\
 \hline
{2D\_Planes} & RMSE & * & 2.163 & 1.4336 & \textbf{1.1251} \\
 & MAE & * & 1.9962 & 1.2799 & 0.9513 \\
 & $\text{Error}_{pos}$ & * & 1.9949 & 1.0785 & 1.003 \\
 & $\text{Error}_{neg}$ & * & 1.9973 & 1.4062 & 0.8978 \\ \hline
{Ailerons} & RMSE & 0.1524 & 0.1587 & \textbf{0.1195} & 0.1365 \\
 & MAE & 0.1461 & 0.1513 & 0.107 & 0.1094 \\
 & $\text{Error}_{pos}$ & 0.1316 & 0.1714 & 0.0963 & 0.1302 \\
 & $\text{Error}_{neg}$ & 0.1654 & 0.1255 & 0.1219 & 0.0426 \\ \hline
{Parkinsons\_Telemonitoring} & RMSE & \textbf{0.0349} & 0.4306 & 0.0449 & 0.0385 \\
 & MAE & 0.0291 & 0.4283 & 0.0379 & 0.0321 \\
 & $\text{Error}_{pos}$ & 0.0206 & - & 0.0398 & 0.033 \\
 & $\text{Error}_{neg}$ & 0.0316 & 0.4283 & 0.0362 & 0.0313 \\ \hline
{Pole\_Telecomm} & RMSE & 1.0588 & 1.1328 & 0.8054 & \textbf{0.5421} \\
 & MAE & 0.9695 & 0.9034 & 0.7546 & 0.4284 \\
 & $\text{Error}_{pos}$ & 1.378 & 1.6409 & 0.8935 & 0.4653 \\
 & $\text{Error}_{neg}$ & 0.7773 & 0.5044 & 0.6924 & 0.408 \\ \hline
 Average RMSE & \multicolumn{1}{l}{} & 0.3113 & 0.6629 & 0.3266 & \textbf{0.2736} \\ \hline
  \multicolumn{6}{l}{$^{\dagger}$ represents the proposed model.}\\
   \multicolumn{6}{l}{The boldface denotes the RMSE of the best model corresponding to each dataset.}\\
   \multicolumn{6}{l}{* denotes that the Matlab program encounters an “out of memory” error.}
\end{tabular}}
\end{table}

\begin{table}[htp]
\centering
\caption{Rank of RMSE for the proposed HE-LSSVR and baseline models on benchmark UCI datasets.}
\label{tab:Rank-table}
\resizebox{\textwidth}{!}{
\begin{tabular}{lcccc}
\hline
Dataset & SVR\cite{drucker1996support} & LS-SVR\cite{suykens1999least} & BLSSVR\cite{fu2023robust} & HE-LSSVR$^{\dagger}$ \\ \hline
2014\_2015 CSM dataset & 3 & 4 & 2 & 1 \\
cpu\_pref & 1 & 4 & 2 & 2 \\
hungary chickenpox & 3 & 4 & 1 & 1 \\
istanbul stock exchange data & 3 & 4 & 2 & 1 \\
mcs\_ds\_edited\_iter\_shuffled & 3 & 4 & 2 & 1 \\
qsar\_aquatic toxicity & 4 & 3 & 2 & 1 \\
Daily\_Demand\_Forecasting\_Orders & 2 & 4 & 3 & 1 \\
bodyfat & 2 & 4 & 3 & 1 \\
machine & 1 & 4 & 2 & 2 \\
slump\_test & 2 & 4 & 3 & 1 \\
Airfoil\_Self\_Noise & 3 & 4 & 2 & 1 \\
Delta\_Ailerons & 3 & 4 & 2 & 1 \\
Forest\_Fires & 3 & 4 & 1 & 1 \\
Yacht\_Hydrodynamics & 3 & 4 & 1 & 2 \\
2D\_Planes & - & 3 & 2 & 1 \\
Ailerons & 3 & 4 & 1 & 2 \\
Parkinsons\_Telemonitoring & 1 & 4 & 3 & 2 \\
Pole\_Telecomm & 3 & 4 & 2 & 1 \\ \hline
Average Rank & 2.5294 & 3.8888 & 2 & 1.2777 \\ \hline
\end{tabular}}
\end{table}

\begin{table}[htp]
\caption{Difference in the RMSE rankings of the proposed HE-LSSVR model against baseline models on UCI datasets.}
\label{tab:Nemenyi-table}
\resizebox{\textwidth}{!}{%
\begin{tabular}{|l|c|c|c|}
\hline
Model & Average rank & Rank difference & \begin{tabular}[c]{@{}c@{}}Significant difference\\ (As per Nemenyi post hoc test)\end{tabular} \\ \hline
SVR \cite{drucker1996support} & 2.5294 & 1.2517 & Yes \\ \hline
LS-SVR \cite{suykens1999least} & 3.8888 & 2.6103 & Yes \\ \hline
BLSSVR \cite{fu2023robust} & 2 & 0.7223 & No \\ \hline
HE-LSSVR$^{\dagger}$ & 1.2777 & - & N/A \\ \hline
\multicolumn{4}{l}{$^{\dagger}$ represents the proposed model.}
\end{tabular}%
}
\end{table}

\begin{table}[htp]
\centering
\caption{Training time in seconds of the proposed HE-LSSVR model and baseline models on benchmark UCI datasets.}
\label{tab:Time-table}
\resizebox{\textwidth}{!}{
\begin{tabular}{lcccc}
\hline
Dataset & SVR\cite{drucker1996support} & LS-SVR\cite{suykens1999least} & BLSSVR\cite{fu2023robust} & HE-LSSVR \\ \hline
2014\_2015 CSM dataset & 0.0119 & 0.0042 & 0.0037 & 0.0027 \\
cpu\_pref & 0.0117 & 0.0046 & 0.0033 & 0.0037 \\
hungary chickenpox & 0.0925 & 0.4889 & 0.0045 & 0.003 \\
istanbul stock exchange data & 0.0657 & 0.0537 & 0.0034 & 0.0031 \\
mcs\_ds\_edited\_iter\_shuffled & 0.004 & 0.002 & 0.0035 & 0.0032 \\
qsar\_aquatic toxicity & 0.0839 & 0.0362 & 0.0037 & 0.0029 \\
Daily\_Demand\_Forecasting\_Orders & 0.0029 & 0.001 & 0.0035 & 0.0031 \\
bodyfat & 0.0136 & 0.0082 & 0.0018 & 0.0044 \\
machine & 0.0135 & 0.0049 & 0.0041 & 0.0039 \\
slump\_test & 0.0069 & 0.0028 & 0.0041 & 0.0033 \\
Airfoil\_Self\_Noise & 0.6113 & 0.3825 & 0.004 & 0.0032 \\
Delta\_Ailerons & 66.606 & 24.4937 & 0.0053 & 0.0074 \\
Forest\_Fires & 0.0616 & 0.0371 & 0.0037 & 0.0028 \\
Yacht\_Hydrodynamics & 0.0177 & 0.0142 & 0.0035 & 0.0112 \\
Ailerons & 152.7565 & 71.5022 & 0.006 & 0.0043 \\
Parkinsons\_Telemonitoring & 24.9768 & 7.6544 & 0.0041 & 0.0103 \\
Pole\_Telecomm & 346.2366 & 95.2581 & 0.0045 & 0.0115 \\ \hline
Average Time & 34.7984 & 11.7617 & 0.0039 & 0.0049 \\ \hline
\end{tabular}}
\end{table}

\subsection{Experiments on Synthetic Datasets}
To further analyze and compare the proposed HE-LSSVR with baseline models, in this subsection we construct synthetic datasets with varying levels of noise. A detailed description of the synthetic datasets used for evaluation is provided in Table \ref{tab:Synthetic datasets description-table}. We have generated five artificial datasets, each of which is duplicated with three different types of noise (Gaussian noise, uniform noise, and student random variable noise), resulting in a total of $15$ artificial datasets, each containing $500$ samples. Table \ref{tab:Synthetic-table} presents the RMSE values for both the proposed HE-LSSVR model and the baseline models on synthetic datasets. Across the 15 synthetic datasets examined, the proposed HE-LSSVR model outperforms the baseline models on 14 datasets, demonstrating its superior performance in the majority of cases. This observation significantly underscores the robustness and effectiveness of the proposed HE-LSSVR model in comparison to the baseline models.

\begin{table}[htp]
\centering
\caption{Description of synthetic datasets.}
\label{tab:Synthetic datasets description-table}
\resizebox{\textwidth}{!}{%
\begin{tabular}{cllc}
\hline
Name                 & Function expression  & $\varrho $                & Variable domain      \\ \hline
                     &  &  & \\
Function 1 & $\quad y_i=\sin  x_i+\varrho_i$ &  & $x_i \in[0,2\pi]$ \\ &   &  &  \\
                      
Function 2 & $\quad y_i=\frac{\sin \left(3 x_i\right)}{3 x_i}+\varrho_i$ &   &  $x_i \in[-4,4]$  \\   &   & $\varrho_i \sim N\left(0,0.2^2\right),$ &  \\

Function 3 & $\quad y_i=\sin x_i \cos x_i^2 +\varrho_i$ &$U[-0.2,0.2],$ & $x_i \in [0,2\pi]$    \\
&  & $T(10)$  & \\

Function 4 & $\quad y_i=x_i \cos x_i+\varrho_i$ &   & $x_i \in[-4,4]$  \\ & & &   
\\
Function 5 & $\quad y_i=(1-x_i+2x_i^2) \exp({\frac{-x_i^2}{2}})+\varrho_i$& & $x_i \in[-4,4]$ \\
\multicolumn{1}{l}{} & \multicolumn{1}{l}{} & \multicolumn{1}{l}{} & \multicolumn{1}{l}{} \\ \hline
\multicolumn{4}{l}{Here, $N\left(0,d^2\right)$ represents the Gaussian random variable with zero mean and standard deviation $d$. }\\
\multicolumn{4}{l}{$U[-a,a]$ represents the uniformly random variable in $[-a,a]$.}\\
\multicolumn{4}{l}{$T(c)$ represents the student random variable with $c$ degree of freedom.}
\end{tabular}%
}
\end{table}

\begin{table}[htp]
\centering
\caption{RMSE values on the synthetic datasets for the proposed HE-LSSVR model against baseline models.}
\label{tab:Synthetic-table}
\resizebox{\textwidth}{!}{
\begin{tabular}{lcrrrc}
\hline
Dataset & Metric & SVR\cite{drucker1996support} & LS-SVR\cite{suykens1999least} & BLSSVR\cite{fu2023robust} & HE-LSSVR \\ \hline
\multicolumn{6}{c}{Type 1: Gaussian noise with 0 mean and 0.2 standard deviation} \\ \hline
Function 1 & RMSE & 0.12381 & 0.70565 & 0.12153 & 9.84$\times$$10^{-6}$ \\
Function 2 & RMSE & 0.0005 & 0.64655 & 0.00459 & 1.93$\times$$10^{-5}$ \\
Function 3 & RMSE & 0.09197 & 0.70557 & 0.12153 & 7.59$\times$$10^{-6}$ \\
Function 4 & RMSE & 0.00127 & 0.64479 & 0.03179 & 8.52$\times$$10^{-5}$ \\
Function 5 & RMSE & 0.00067 & 0.6585 & 0.0296 & 7.34$\times$$10^{-5}$ \\ \hline
\multicolumn{6}{c}{Type 2: Uniform noise over the interval $\left[-0.2, 0.2\right]$} \\ \hline
Function 1 & RMSE & 0.07732 & 0.70559 & 0.12114 & 9.84$\times$$10^{-6}$ \\
Function 2 & RMSE & 0.00023 & 0.64655 & 0.00459 & 1.93$\times$$10^{-5}$ \\
Function 3 & RMSE & 0.0849 & 0.70555 & 0.12153 & 7.59$\times$$10^{-6}$ \\
Function 4 & RMSE & 0.00101 & 0.64162 & 0.01218 & 1.58$\times$$10^{-5}$ \\
Function 5 & RMSE & 0.00061 & 0.66018 & 0.0296 & 7.27$\times$$10^{-5}$ \\ \hline
\multicolumn{6}{c}{Type 3: Student random variable noise with 10 degrees of freedom} \\ \hline
Function 1 & RMSE & 0.10258 & 0.7056 & 0.12153 & 4.1 $\times$ $10^{-6}$ \\
Function 2 & RMSE & 0.00103 & 0.64677 & 0.00459 & 5.81$\times$$10^{-5}$ \\
Function 3 & RMSE & 0.00174 & 0.68897 & 0.00319 & 2.74$\times$ $10^{-7}$ \\
Function 4 & RMSE & 0.00121 & 0.64258 & 0.03179 & 1.42$\times$$10^{-5}$ \\
Function 5 & RMSE & 0.00018 & 0.6595 & 0.0296 & 0.00019 \\ \hline
\end{tabular}}
\end{table}

\subsection{Experiments on Time Series Datasets}
To validate the superiority of the proposed HE-LSSVR model in real-world applications, we further evaluate it on time series datasets \cite{derrac2015keel}. The experimental results of the proposed HE-LSSVR and the baseline models are presented in Table \ref{tab:Timeseries-table}. The proposed HE-LSSVR showcases remarkable performance as it has the lowest RMSE value on each of the $11$ time series datasets. The average RMSE values of the proposed HE-LSSVR and the baseline models (SVR, LS-SVR, and BLSSVR) are $2.2772$, $2.6727$, $5.2239$, and $2.5178$, respectively. These values represent the overall performance of each model, with lower RMSE values indicating better predictive accuracy. The results strongly reflect the prominence of the proposed HE-LSSVR model compared to the baseline models.
\begin{table*}[htp]
\centering
\caption{RMSE, MAE, $\text{Error}_{pos}$, and $\text{Error}_{neg}$ values on the time series datasets for the proposed HE-LSSVR model against baseline models. }
\label{tab:Timeseries-table}
\resizebox{\textwidth}{!}{
\begin{tabular}{cccccc}
\hline
\begin{tabular}{@{}c@{}}Dataset\\ (No. of samples, No. of features)\end{tabular} & Metric & SVR\cite{drucker1996support} & LS-SVR\cite{suykens1999least} & BLSSVR\cite{fu2023robust} & HE-LSSVR$^{\dagger}$ \\ \hline
{\begin{tabular}[c]{@{}c@{}}NNGC1\_dataset\_D1\_V1\_003\\ (430, 5)\end{tabular}} & RMSE & 0.6837 & 0.7501 & 0.6192 & \textbf{0.5369} \\
 & MAE & 0.5544 & 0.593 & 0.5043 & 0.4444 \\
 & $\text{Error}_{pos}$ & 0.5957 & 0.6757 & 0.5316 & 0.4377 \\
 & $\text{Error}_{neg}$ & 0.5168 & 0.4536 & 0.4715 & 0.4526 \\ \hline
{\begin{tabular}[c]{@{}c@{}}NNGC1\_dataset\_D1\_V1\_004\\ (545, 5)\end{tabular}} & RMSE & 0.562 & 0.6041 & 0.4883 & \textbf{0.4221} \\
 & MAE & 0.4701 & 0.5045 & 0.4047 & 0.3459 \\
 & $\text{Error}_{pos}$ & 0.2759 & 0.2825 & 0.4481 & 0.3219 \\
 & $\text{Error}_{neg}$ & 0.5699 & 0.5812 & 0.3764 & 0.3627 \\ \hline
{\begin{tabular}[c]{@{}c@{}}NNGC1\_dataset\_D1\_V1\_005\\ (430, 5)\end{tabular}} & RMSE & 0.0307 & 0.4047 & 0.0328 & \textbf{0.0298} \\
 & MAE & 0.0255 & 0.4028 & 0.0274 & 0.0247 \\
 & $\text{Error}_{pos}$ & 0.025 & 0.4028 & 0.0222 & 0.0252 \\
 & $\text{Error}_{neg}$ & 0.0259 & - & 0.0301 & 0.0243 \\ \hline
{\begin{tabular}[c]{@{}c@{}}NNGC1\_dataset\_D1\_V1\_006\\ (610, 5)\end{tabular}} & RMSE & 0.1697 & 0.4305 & 0.1678 & \textbf{0.1566} \\
 & MAE & 0.138 & 0.3945 & 0.136 & 0.128 \\
 & $\text{Error}_{pos}$ & 0.1426 & 0.3967 & 0.1532 & 0.1283 \\
 & $\text{Error}_{neg}$ & 0.1337 & 0.1237 & 0.1136 & 0.1277 \\ \hline
{\begin{tabular}[c]{@{}c@{}}NNGC1\_dataset\_D1\_V1\_007\\ (610, 5)\end{tabular}} & RMSE & 0.086 & 0.4104 & 0.0877 & \textbf{0.0833} \\
 & MAE & 0.0673 & 0.4013 & 0.069 & 0.065 \\
 & $\text{Error}_{pos}$ & 0.0679 & 0.4013 & 0.0639 & 0.0662 \\
 & $\text{Error}_{neg}$ & 0.0668 & - & 0.0724 & 0.0638 \\ \hline
{\begin{tabular}[c]{@{}c@{}}NNGC1\_dataset\_D1\_V1\_008\\ (540, 5)\end{tabular}} & RMSE & 0.0982 & 0.4195 & 0.0998 & \textbf{0.0949} \\
 & MAE & 0.0773 & 0.4078 & 0.0762 & 0.075 \\
 & $\text{Error}_{pos}$ & 0.0693 & 0.4078 & 0.0585 & 0.0718 \\
 & $\text{Error}_{neg}$ & 0.0868 & - & 0.0939 & 0.0789 \\ \hline
{\begin{tabular}[c]{@{}c@{}}NNGC1\_dataset\_D1\_V1\_009\\ (540, 5)\end{tabular}} & RMSE & 0.2481 & 0.44 & 0.2368 & \textbf{0.2256} \\
 & MAE & 0.1938 & 0.3795 & 0.1866 & 0.1771 \\
 & $\text{Error}_{pos}$ & 0.1834 & 0.4052 & 0.1905 & 0.1746 \\
 & $\text{Error}_{neg}$ & 0.2055 & 0.0972 & 0.1808 & 0.179 \\ \hline
{\begin{tabular}[c]{@{}c@{}}NNGC1\_dataset\_D1\_V1\_010\\ (585, 5)\end{tabular}} & RMSE & 0.0908 & 0.3931 & 0.0939 & \textbf{0.0849} \\
 & MAE & 0.071 & 0.3781 & 0.0737 & 0.0692 \\
 & $\text{Error}_{pos}$ & 0.073 & 0.3781 & 0.0607 & 0.0724 \\
 & $\text{Error}_{neg}$ & 0.069 & - & 0.0827 & 0.0625 \\ \hline
 \end{tabular}}
\end{table*}

\begin{table*}[htp]
\centering
\caption{(Continued) RMSE, MAE, $\text{Error}_{pos}$, and $\text{Error}_{neg}$ values on the time series datasets for the proposed HE-LSSVR model against baseline models. }
\label{tab:Timeseries-table}
\resizebox{\textwidth}{!}{
\begin{tabular}{cccccc}
\hline
\begin{tabular}{@{}c@{}}Dataset\\ (No. of samples, No. of features)\end{tabular} & Metric & SVR\cite{drucker1996support} & LS-SVR\cite{suykens1999least} & BLSSVR\cite{fu2023robust} & HE-LSSVR$^{\dagger}$ \\ \hline
{\begin{tabular}[c]{@{}c@{}}NNGC1\_dataset\_E1\_V1\_008\\ (740, 5)\end{tabular}} & RMSE & 0.2226 & 0.4569 & 0.2182 & \textbf{0.2122} \\
 & MAE & 0.1842 & 0.4083 & 0.1773 & 0.1711 \\
 & $\text{Error}_{pos}$ & 0.1644 & 0.4374 & 0.147 & 0.1251 \\
 & $\text{Error}_{neg}$ & 0.2103 & 0.1065 & 0.2194 & 0.2512 \\ \hline
{\begin{tabular}[c]{@{}c@{}}NNGC1\_dataset\_E1\_V1\_009\\ (740, 5)\end{tabular}} & RMSE & 0.2132 & 0.4538 & 0.207 & \textbf{0.1995} \\
 & MAE & 0.1778 & 0.4076 & 0.1641 & 0.1677 \\
 & $\text{Error}_{pos}$ & 0.1563 & 0.4267 & 0.1171 & 0.1325 \\
 & $\text{Error}_{neg}$ & 0.2068 & 0.1131 & 0.2293 & 0.2347 \\ \hline
{\begin{tabular}[c]{@{}c@{}}NNGC1\_dataset\_E1\_V1\_010\\ (650, 5)\end{tabular}} & RMSE & 0.2676 & 0.4607 & 0.2662 & \textbf{0.2315} \\
 & MAE & 0.2139 & 0.4073 & 0.2167 & 0.1661 \\
 & $\text{Error}_{pos}$ & 0.1866 & 0.4399 & 0.2139 & 0.1329 \\
 & $\text{Error}_{neg}$ & 0.2457 & 0.137 & 0.2238 & 0.2003 \\ \hline
 Average RMSE & \multicolumn{1}{l}{} & 0.243 & 0.4749 & 0.229 & \textbf{0.207} \\ \hline
 \multicolumn{6}{l}{$^{\dagger}$ represents the proposed model.}\\
   \multicolumn{6}{l}{The boldface denotes the RMSE of the best model corresponding to each dataset.}
   
\end{tabular}}
\end{table*}

\section{Conclusions and Future work}
In this paper, we addressed the limitations of support vector regression (SVR) by introducing the HawkEye loss function, a novel symmetric loss function that is bounded, smooth, and simultaneously possesses an insensitive zone. We presented a comparative analysis of existing loss functions and demonstrated that the HawkEye loss function is the first of its kind in the SVR literature. Further, we integrated the HawkEye loss function into the least squares framework of SVR and developed a new fast and robust model called HE-LSSVR. To solve the optimization problem of HE-LSSVR, we employed the adaptive moment estimation (Adam) algorithm, which offers adaptive learning rates and is effective in handling large-scale problems. This is the first time Adam has been utilized to solve an SVR problem. Through extensive numerical experiments on UCI, synthetic, and time series datasets, we empirically validated the superiority of the proposed HE-LSSVR model. It exhibits remarkable generalization performance and efficiency in training time compared to baseline models. These findings collectively positioned the HE-LSSVR model as a superior choice for real-world applications in diverse fields, affirming its significance and potential impact in the realm of regression tasks.
\par
In the future, there is a potential avenue for delving into the integration of the HawkEye loss function within the domain of deep learning models. The essential characteristics of boundedness and smoothness intrinsic to the HawkEye loss function suggest promising possibilities for its fusion with advanced machine learning and deep learning techniques. This future direction holds the potential to yield substantial advancements and broader applicability across diverse domains.

\section*{Acknowledgment}
This project is supported by the Indian government's Science and Engineering Research Board (MTR/2021/000787) under the Mathematical Research Impact-Centric Support (MATRICS) scheme. The Council of Scientific and Industrial Research (CSIR), New Delhi, provided a fellowship for Mushir Akhtar’s research under grant no.
09/1022(13849)/2022-EMR-I.

\bibliography{refs.bib}
\bibliographystyle{unsrtnat}
\end{document}